Kais HADDAR, Hela FEHRI, Laurent ROMARY

# A prototype for projecting HPSG syntactic lexica towards LMF

## Abstract

The comparative evaluation of Arabic HPSG grammar lexica requires a deep study of their linguistic coverage. The complexity of this task results mainly from the heterogeneity of the descriptive components within those lexica (underlying linguistic resources and different data categories, for example). It is therefore essential to define more homogeneous representations, which in turn will enable us to compare them and eventually merge them.

In this context, we present a method for comparing HPSG lexica based on a rule system. This method is implemented within a prototype for the projection from Arabic HPSG to a normalised pivot language compliant with LMF (ISO 24613 - Lexical Markup Framework) and serialised using a TEI (Text Encoding Initiative) based representation. The design of this system is based on an initial study of the HPSG formalism looking at its adequacy for the representation of Arabic, and from this, we identify the appropriate feature structures corresponding to each Arabic lexical category and their possible LMF counterparts.

## 1 INTRODUCTION

HPSG (Head-driven Phrase Structure Grammar) syntactic lexica have been developed as part of various applications such as parsing of natural language and construction of electronic dictionaries (Blache, 1995; Levine and Meurers, 2006; Pollard and Sag, 1994). The evaluation, reclaim and exploitation of the results provided by these applications are often seen as complex tasks because they are generally not based on normalised lexical resources. Additionally, the corresponding lexical resources are not described on the basis of the same underlying descriptors (or "data categories", to use the terminology of ISO 12620:2009 - see Ide and Romary, 2004). It is therefore important to define a conceptual framework that allows the definition of a pivot language between such resources in order to construct normalised representations from existing ones using merging and interoperability mechanisms. In line with the principles articulated in (Romary and Ide, 2004), the pivot language should be based on a standardised abstract meta-model combined with data categories. This in turn makes it possible to implement the pivot language using any kind of concrete syntax, i.e. an XML vocabulary, that maps onto the abstract model in an isomorphic way.

This paper follows these modelling principles with the main objective of proposing a method for the transformation of HPSG grammar lexica into a normalised pivot language that conforms to the principles of the LMF standard (ISO 24613), a framework that has been designed by the ISO committee TC 37/SC 4. More specifically, the pivot language will be used to estimate the real coverage of existing HPSG syntactic lexica and to merge them into integrated resources. It is worth noting that the same process can also be applied to lexica defined under other unification formalisms.

The proposed method takes into account both the specificities of the HPSG formalism as adapted to the Arabic language and the possibility of applying LMF to this formalism. This paper accordingly provides a precise overview of both formalisms with a view to identifying the adaptations that can be brought to HPSG and the data categories that must be added to LMF in accordance with ISO standard 12620. This study will enable us to elaborate a rule-based system for projecting HPSG syntactic lexicons towards LMF in a systematic way.

Section 2 of this paper presents the main reference works that have either covered standardisation attempts in the language resource domain or actual methods for projecting information across formalisms. We then briefly present in section 3 the HPSG formalism and the linguistic phenomena that may be covered by this formalism. We subsequently introduce in section 4 the LMF platform and its main principles. Section 5 focuses on our method of projecting an HPSG grammar lexicon for the Arabic language towards LMF as well as the experimentation of this method. We conclude on possible further ways this work could be extended to other types of lexical resources.

## 2 NORMALISATION AND PROJECTION ACTIVITIES

There have been several works dealing with the use of HPSG lexica for the processing of the Arabic language, including (Abdelkader, 2006), (Chabchoub, 2005), and (Elleuch, 2004). Still, the corresponding lexical resources are small and each of them concentrates on a particular task or syntactic phenomenon. Despite this, when considered together, they are highly complementary and merging them could definitely lead to a much richer lexicon containing a wide variety of lexical categories for the Arabic language. The fusion operation is, however, quite complex. The HPSG formalism can be implemented in different ways depending on the underlying theoretical assumptions as well as the actual language being dealt with. For instance, some features can be found in one lexicon but not in another one depending on the underlying linguistic viewpoint. For example, a feature like /slash/ will only appear in the context of elliptical or relative constructs. Additionally, the actual technical implementation when computerised may come in various organisations (e.g. feature granularity) and formats (e.g. XML, binary). This makes the recovery of the corresponding lexical content from one application to another extremely complex.

In order to achieve the reuse of such resources as well as their fusion, on the basis of standardised data categories, it is necessary to adopt a comprehensive normalisation strategy. To this end, research has been continuously carried out in recent years (see, for example, Ide and Véronis, 1995; Monachini and Calzolari, 1999; Atkins *et al.*, 2002) so that a community of researchers using a given formalism can benefit from the results, lexicons and resources developed by other communities using various formalisms. These endeavours have taken a range of dictionary models as a basis and suggested lexical abstraction adapted to automatic language processing, and at the same time has sought to retain the best compromise between simplicity and wide coverage. Specific attempts (see Eagles, 1996, for an example of a cross-formalism survey) have been made to standardise under-categorisation processes. They rely on a comparison between linguistic formalisms and NLP lexicons so that it is possible to carry out transformations from one formalism into another.

The continuous stream of projects and activities such as *GENELEX* (Genelex, 1994), *EAGLES*, *ISLE*, *MULTEXT*, *TEI* (Lemnitzer *et al.*, to appear), together with the mass of expertise that these have contributed to, has led to the finalisation of an ISO standard on the representation of computerised lexical structures, namely ISO 24613 LMF within ISO committee TC 37/SC 4[1]. Published in 2008, this initiative has already been followed up by several attempts to provide reference implementations compliant with the future standard. In the domain of morphological lexica, for instance, the *Morphalou* project (Romary *et al.*, 2004) provides a full-form lexicon for French comprising 540,000 inflected forms. Nguyen and colleagues (Nguyen *et al.*, 2006) also describe the implementation of LMF for a full-featured lexicon for NLP purposes. A morphological lexicon, *ArabicLDB* (Khemakhem 2006), has been proposed for Arabic: it exemplifies in particular how to implement roots and vocalic patterns for Arabic morphology.

Similarly, several tools have already been proposed to help construct lexical databases in conformity with LMF together with standardised data categories. In particular, Lexus (Ringersma & Kemps-Snijders, 2007) is an online environment allowing one to both model a lexical structure compliant with the LMF meta-model and import lexical content accordingly. An endeavour to develop an editor with a constraint checker for the Arabic language has recently been proposed (Hasni *et al.*, 2006).

From the point of view of cross-formalism mapping, we can identify two main trends. The first one corresponds to the simplified use of specialised concept lexica. It presents the risk of getting ill-formed structures during analysis because it does not take into account specificities of each formalism. The second approach uses a rule-based system that takes the role of a parser. This approach is more efficient than the first as it always yields well-formed representations. For example, the method presented in (Kasper *et al.*, 1995) translates an HPSG grammar into a TAG (Tree Adjoining Grammar) representation. The underlying translator implements an algorithm that fulfils TAG specific constraints. These constraints define the mapping between the concepts used in the two formalisms. Conversely, in (Yoshinaga *et al.*, 2002), the authors propose an algorithm for converting LTAG into HPSG. While these formalisms treat the same set of constraints, the algorithm consists of mapping the constraints of LTAG one by one into HPSG equivalent ones.

The problem of evaluating and comparing grammars is treated by (Fehri *et al.*, 2006) and (Loukil, 2006). The proposed solution uses LMF as a pivot language and translates the input lexica into an LMF compliant structure.

The knowledge database DIINAR.1[2] encompasses 19,457 verbs, 70,702 deverbal entries, verbal nouns, active and passive participles, 'analogous' adjectives, nouns 'of time and operating place', 39,099 nominal stems, 445 tool-words and a prototype of 1,384 proper names. From this database, a large lexicon can be generated. As an application this database is associated with a morphological analyser called AraParse. This analyser uses a large stem-based lexicon generated form DIINAR.1.

Every entry is associated with morpho-syntactic specifiers at word-level and ensuring grammar-lexis relations between the lexical basis of a given word-form and other word-

1 http://www.iso.org/iso/home/store/catalogue_tc/catalogue_tc_browse.htm?commid=297592

2 DIctionnaire INformatisé de l'ARabe version 1 (see http://silat.univ-lyon2.fr/Presentation%20DIINAR.html)

formatives. The total amount of minimal words (e.g. of lemmas with their prefixes and suffixes) generated from the database is 7,774,938.

The AraParse lexicon contains:

− all the 121,522 unvocalised stem-entries of the DIINAR.1 database,

− all the vocalic schemes of each stem,

− all possible combinations of (prefixes, suffixes) for each couple of stem and vocalic schemes, and a set of specifiers (Genelex, 1994) containing morpho-syntactic information,

− a specifier of compatibility with possible clitics for each triple of stem, vocalic scheme, prefixes/suffixes combination.

The lexicon is organised in a letter tree structure. The principal advantage of the tree structure is that it greatly facilitates access while at the same time considerably reducing the size of the lexicon.

## 3 HPSG ARABIC LEXICON

HPSG has been proposed since the beginning of the eighties by Pollard and Sag (Pollard and Sag, 1994). It belongs to a family of formalisms based on constraints and descends from other previous unification formalisms such as GPSG (Generalised Phrase Structure Grammar), CG (Categorical Grammar) and LFG (Lexical Functional Grammar). It was initially designed to represent Romance or Germanic languages but, in the last decade, has been extensively applied to a wide variety of other language families.

In view of the specificities of the Arabic language (e.g. hierarchy types and agreement), its representation in HPSG requires the modification of some of HPSG's feature values and the addition of some Arabic specific features. To illustrate these specificities, we provide some examples of core features that are routinely needed in the context of the linguistic description of Arabic:

- **CFORM:** this feature is used for the description of the consonantal pattern of verbs and can take one of the values triliteral *thulāthī,* three consonant root or quadriliteral *rubāʿī,* four consonant root. This feature is necessary to identify different schemas that are useful for referencing the canonical or derivative form of the concerned lexical entry (e.g. *ktb* is *thulāthī* and *zqzq* is *rubāʿī*).
- **DENUDE:** is used for triliteral verbs and can take one of the values denuded *mujarrid*, when no extra letters are combined to the root, or increased *mazīd* when the root is combined with extra letters (e.g. *kataba* [wrote] is *mujarrid* and *'inkataba* [was written] is *mazīd*). It is useful in the same contexts as CFORM.
- **DIMINUTIVE:** can take one of the values non diminutive *ghair muṣaghar* or diminutive *ṣīghit al-ttaṣghīr*. With this feature we can distinguish canonical forms from inflected ones (e.g. *kalb* [dog], *kulayeb* [small dog]).
- **RELATIVE:** this feature has the same role as DIMINUTIVE. It can take one of the values relative *manṣūb* or non relative *ghair manṣūb* (e.g. *tūnisī* [Tunisian] is *manṣūb*).
- **NATURE:** is used to give the semantic role of a noun (e.g. 'gift' vs. 'giver'). Among the values that can be taken by this feature, we have: agent noun, *ism fāʿil* (e.g. kātib [writer]), patient noun, *ism mafʿul* (e.g. *maktūb* [written]), verbal adjective, *ṣifa muchabbaha* (e.g. *shujāʿ* [courageous]). Every value taken by this feature represents a lexical entry and a derived or inflected form.

- **RADICAL:** This feature gives the root of a verb (e.g. the root of *kataba* is *ktb*).

We have also modified the definition of the feature NFORM. NFORM gives the different forms that an Arabic noun can have. The values of this feature are: *mutaṣṣarf muchtak* (inflectional derivative), *mutaṣṣarf jāmed* (inflectional inert) and *ghair mutaṣṣarf* (non inflectional). With this feature we can know if a canonical form can have inflected (or even derived) forms or not.

As in (Dahdah, 1992), we consider that an Arabic word can be a noun *ism*, a particle *harf* or a verb *fiʿl*. We can mention here that other categorisations have been used, which usually add the adjective as a fourth category. In what follows we are going to give a preview on the considered word categories.

### 3.1 Nouns

What distinguishes the Arabic language from other languages is the fact that the lexical category for a noun can be broken down into several subcategories to distinguish between frozen (e.g. proper nouns *asmāʾ al-ʿalam*, place nouns *asmāʾ al-makān*), non-frozen (e.g. adjectives *al-ṣṣifa al-muchabbaha*, noun-agent *ism elfāʿil*) and inflected nouns (e.g. demonstrative pronouns *asmāʾ al-ichāra*, pronouns *al-ḍamāʾir*).

Note that all nouns share the same AVM (Attribute Value Matrix) model, represented in Figure 1, where they only differ from one another depending on associated feature values. In the case of adjectives, we add to this skeleton the feature MOD in feature HEAD. The feature MAJ is used to introduce the lexical category of a word (e.g. verb, noun and preposition). The feature DEFN gives the noun the property of definiteness.

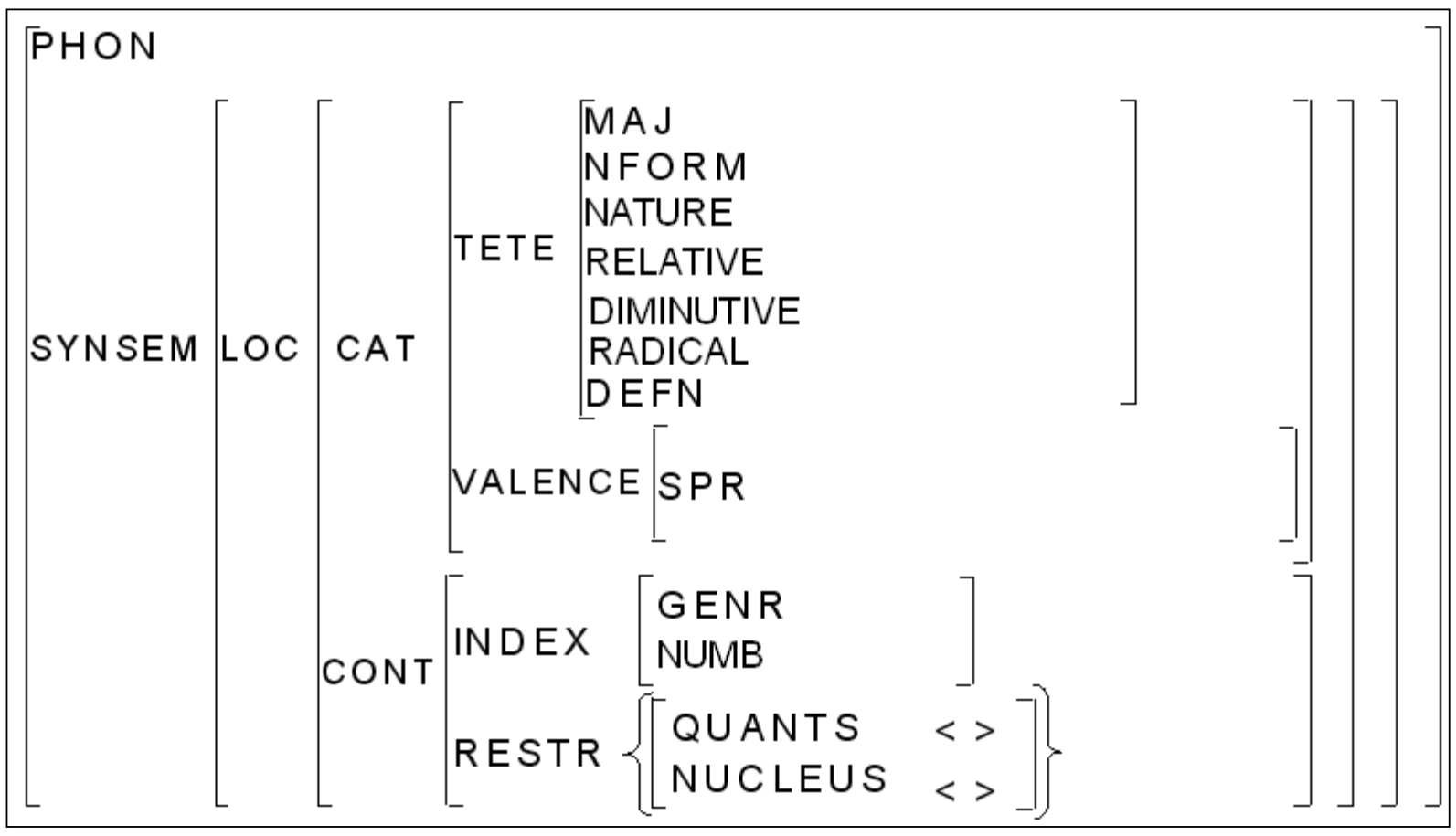

**Figure 1:** A noun AVM model

In a noun AVM, all morphological features are regrouped in the HEAD feature. The only syntactic feature is SPR. This feature introduces the element that precedes a word (e.g. a demonstrative pronoun is a potential value of a noun SPR). Although agreement features are considered as semantic features in HPSG, they are founded in the morphological part in LMF. For each subcategory of the category noun, we must specify an adequate AVM, for example:

**Inert variable noun (*elism elmutassaref eljamed*):** concrete noun' *ism al-thāt*, or abstract noun' *ism al-maʿna*. Figure 2 and Figure 3 are two examples of a concrete noun and an abstract noun.

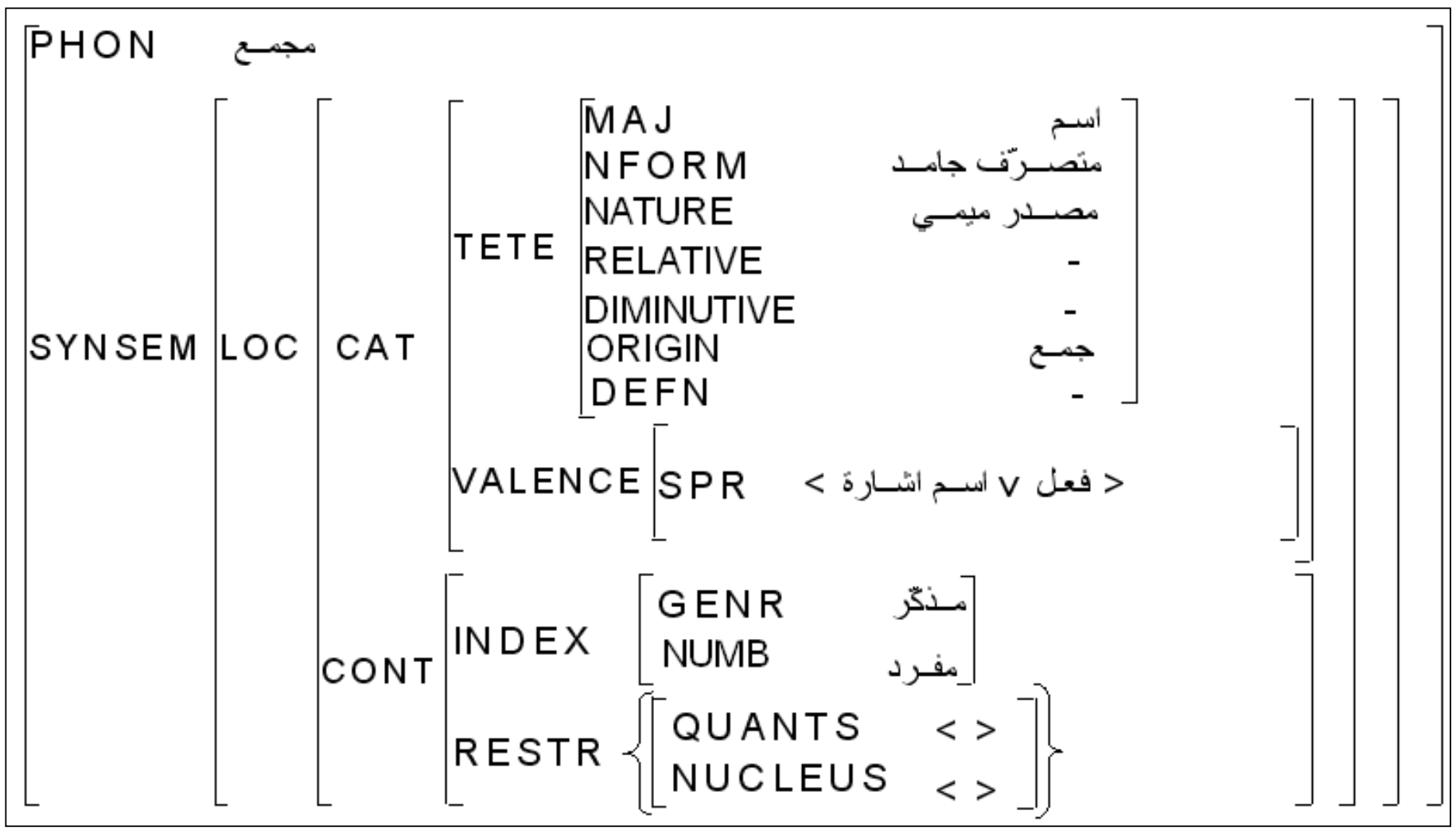

**Figure 2:** AVM of the noun *majmaʿ* (collector**)**

Figure 2 shows that the noun *majma'* (collector) is an inert indefinite noun, non diminutive and non relative. This information is given respectively by the features NFORM, DEFN, DIMINUTIVE and RELATIVE. *Almajma'* is the definite form of *majma'*. The features NATURE, ROOT and SPR show that this noun represents an m-initial infinitive *maṣdar* mīmī having like root *jm'* and that it can be preceded by a verb or a demonstrative pronoun. Note that the noun *majma'* (collector) is a masculine noun and singular.

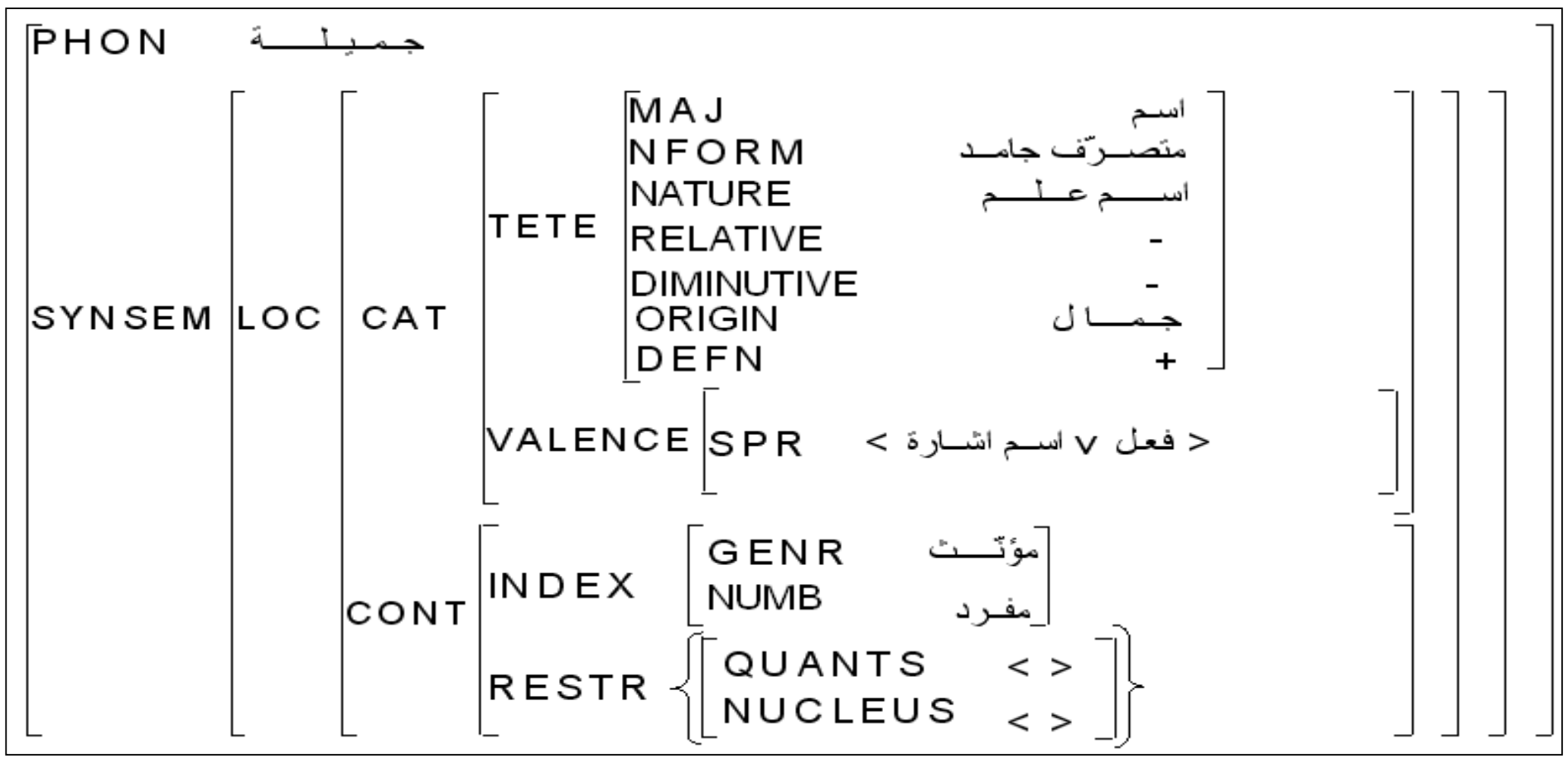

**Figure 3:** AVM of the proper noun *jamīla*

In Figure 3, we notice that the features that changed value are the features NATURE, DEFN ORIGIN, and GENR since *jamīla* is a proper noun, definite and feminine. The *masdar* (ORIGIN) value of the proper noun *jamīla* is *jamāl*. This noun can be preceded by a verb or by a demonstrative pronoun (the SPR value). In Arabic, a proper noun can be used as an adjective and in this case it is necessary to apply some modifications to the appropriate AVM.

### 3.2 Particles

Particles are words that serve to situate events and objects in relation to time and to space. They give a text a coherent sequence. Particles represent another category for an Arabic word and can be construction letters *hurūf mabān* or significance letters *hurūf mabān*. Significance letters are divided into two subcategories: the first regroups particles that have no effect (e.g. morphological, grammatical) on the word whereas the second includes particles that have some declination effects on the noun (e.g. prepositions, particles of the vocative) or on the verb (e.g. elision particles, subjunctive particles) or on both (e.g. conjunctions).

For particles, we proposed two different AVM models; one is used for prepositions and the other for particles. Both AVM models are illustrated respectively in Figure 4 and Figure 5.

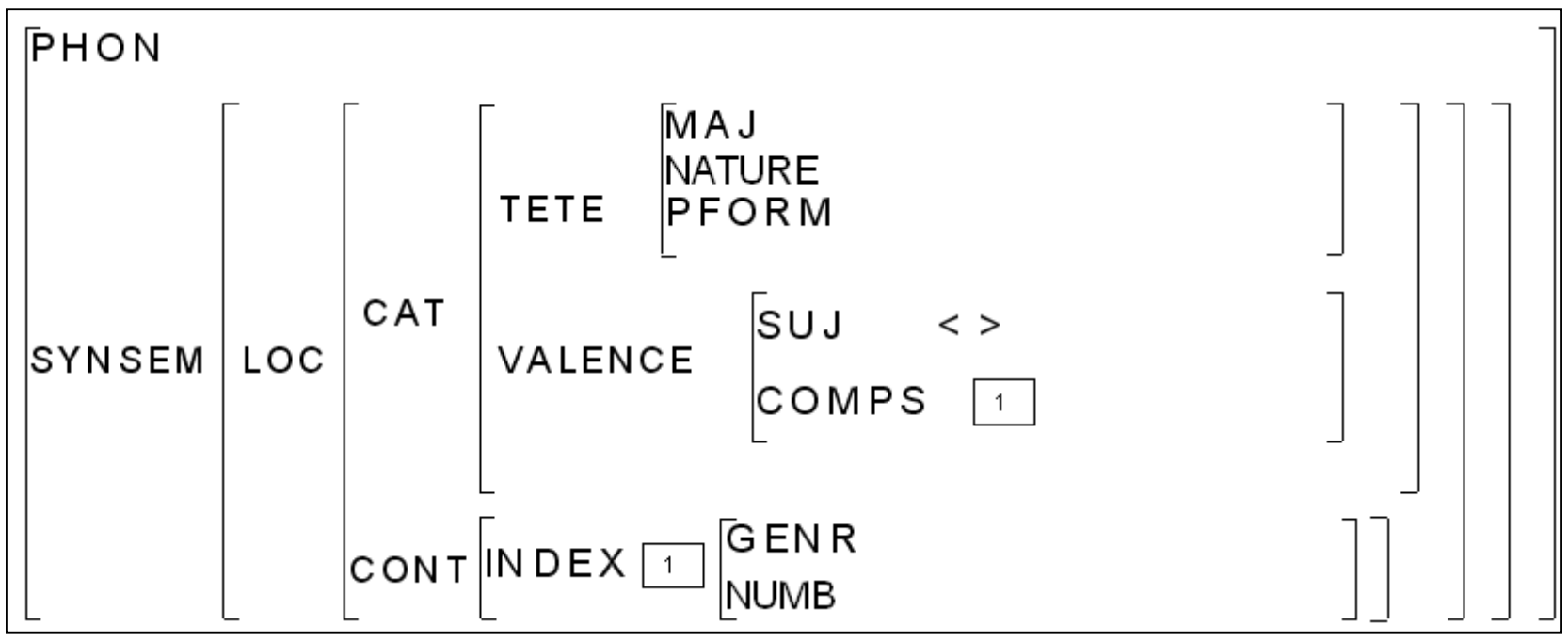

**Figure 4:** AVM model of a preposition

Preposition morphological features are regrouped in the HEAD feature. Note that agreement features are relative to the object introduced by this preposition.

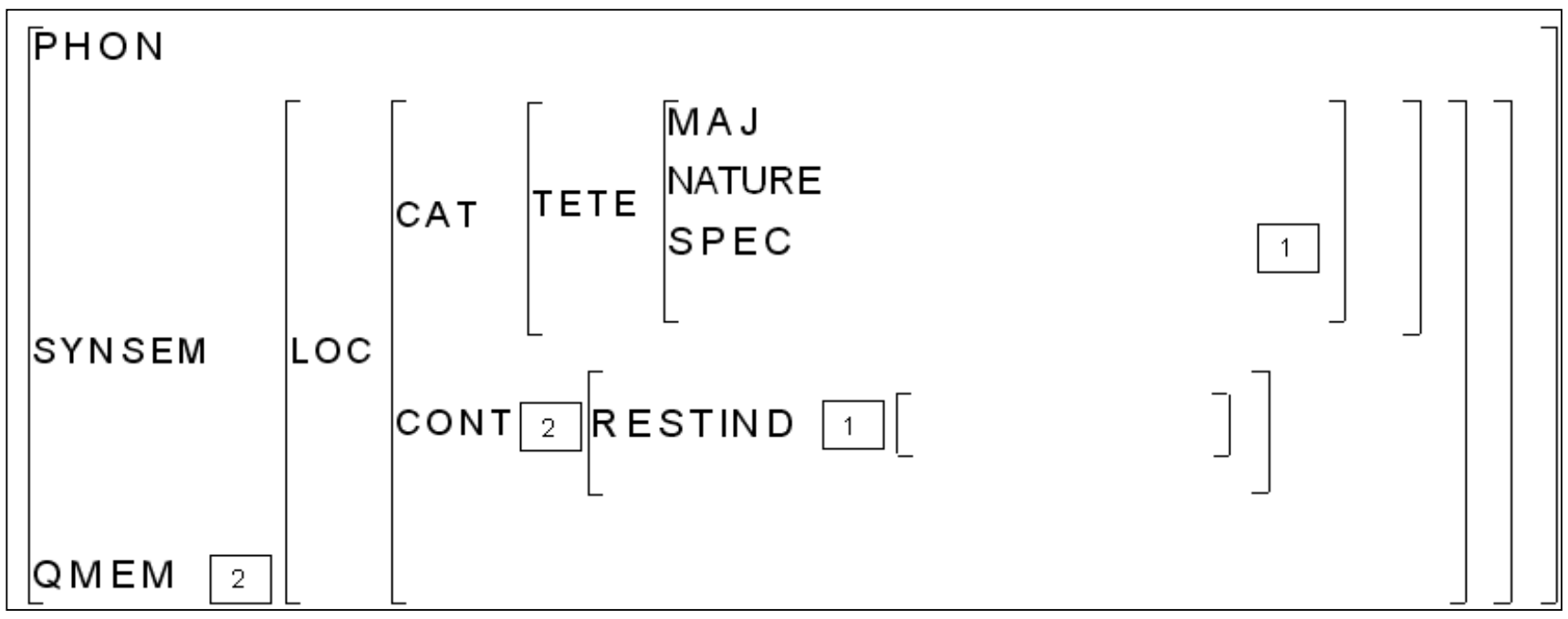

**Figure 5:** AVM model of an elision or subjunctive particle

Note that a preposition AVM is different from that of an elision or subjunctive particle. Elision and subjunctive particles are words that can precede verbs. This difference resides in the features HEAD, VALENCE and CONT. In the preposition AVM we remark the existence of the feature PFORM as a morphological feature. In the tool AVM, the feature SPEC replaces it. Additionally, the features VALENCE and INDEX and agreement features exist for a preposition but not for a tool. In the following figures we are going to give some examples of AVMs that correspond to different categories of particles.

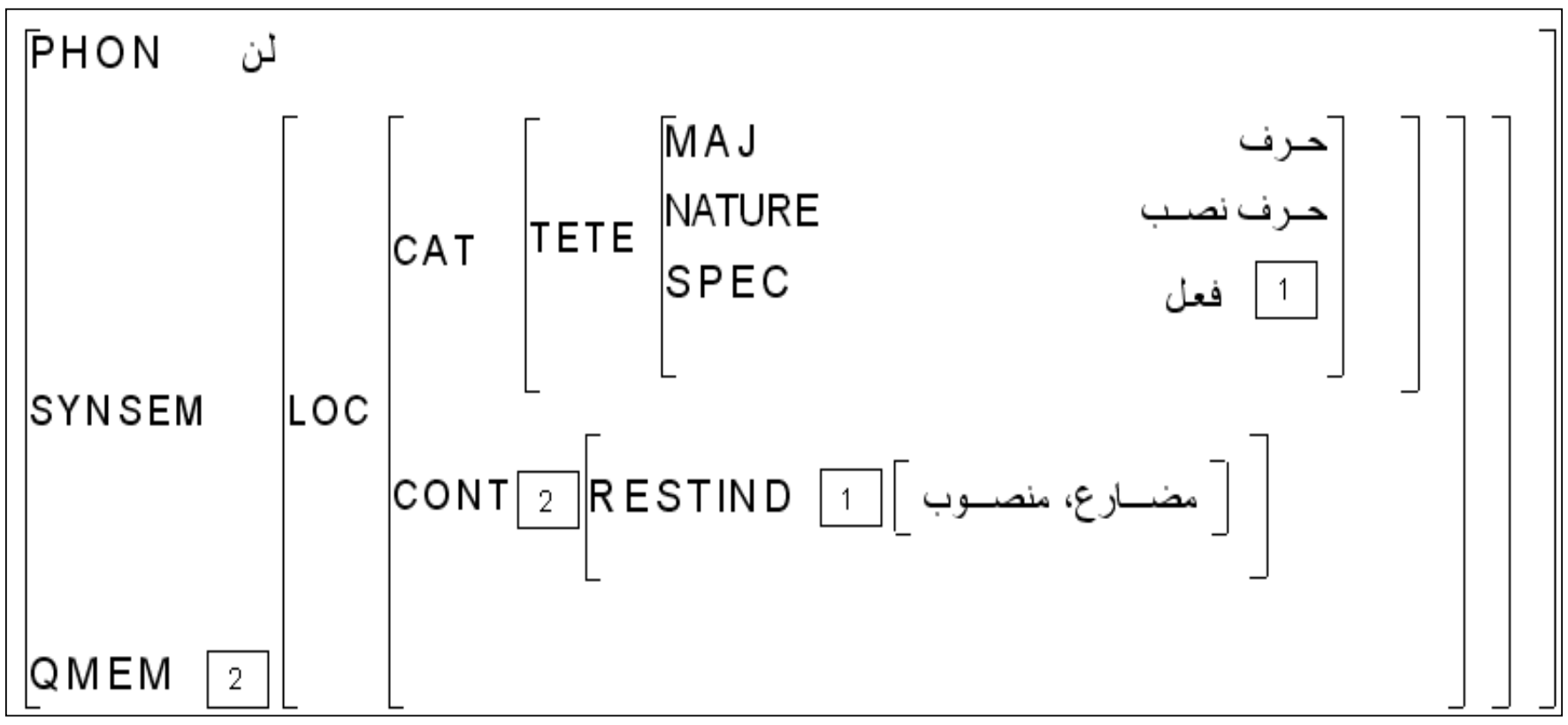

**Figure 6:** AVM of the particle *lan*

Figure 6 represents the tool AVM of the Arabic word *lan*. This word is a particle (value of MAJ) that belongs to significance letters (value of NATURE) and that precedes a verb (SPEC) in the subjunctive mood *manṣub*. The verb must be conjugated in imperfect tense *mudhāra*ʿ (value of RESTIND).

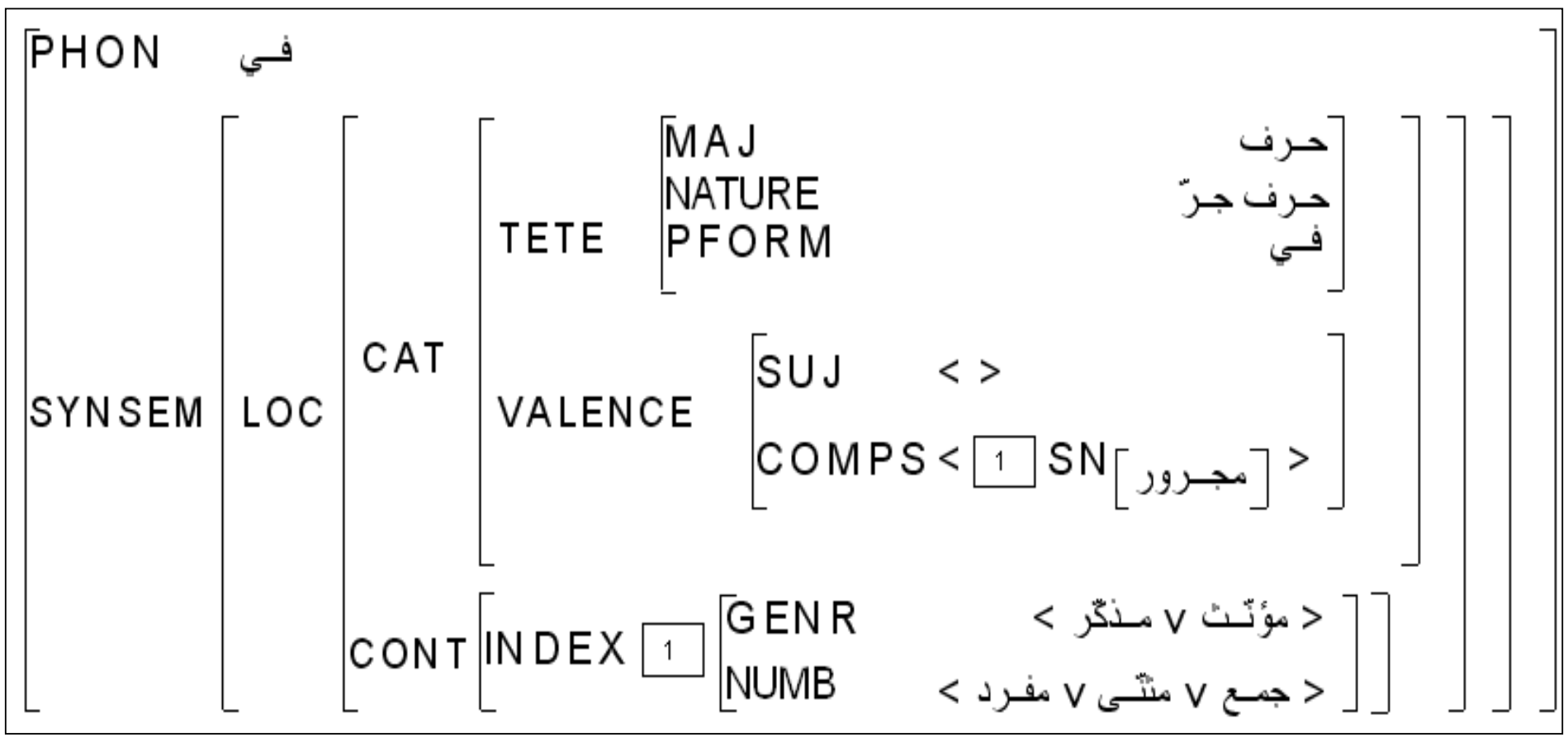

**Figure 7:** AVM of the particle *fī*

Figure 7 represents an example of a preposition AVM. In this AVM we note the existence of the feature VALENCE. The particle *fī* admits an object described by the feature COMPS. This object must be a genitive nominal phrase *majrūr*. The agreements of this object are expressed in the feature INDEX.

### 3.3 Verbs

A verb usually indicates a real action on the part of the subject that occurs over a period of time (e.g. *kataba* [wrote] and *qara'* [read]). It is a fundamental element to which the sentence constituents are connected directly or indirectly. In Arabic, the basic source of all the forms of a verb is called the root of the verb. The root is not a real word; rather it is a sequence of three consonants that can be found in all the words that are related to it. Most roots are composed of three letters, a very few are composed of four or five letters. The verb is therefore the stem of a word family (Ammar and Dichy, 1999).

Schemes are applicable to roots and these applications produce a new verb. For example, from the root *kharaja*, meaning "to go out", we obtain the verb "to make go out" by doubling the central consonant to make *kharraja*. The scheme can be considered as a formal representation established by three or four consonants *fʿl* that are totally vocalised, or as a mould containing the root. Altogether there are 19 verbal schemes that can be either nude[3] or increased by taking three consonants from the root and modifying the vowels, redoubling the second letter of the root, or inserting affixes (prefix, infix and suffix). The longer verbs conjugate with the same prefixes and suffixes as the original verb. Therefore, a root can generate most of the 19 verbs and the corresponding schemes can give 22 different conjugation patterns. In fact, there is a scheme *fa'ala* that can have three variations different from conjugation according to the nature of the vowel used in the second consonant of the root: *yaf' ulu*, *yaf' ilu*, and *yaf' alu*. Also, the scheme *fa 'ila* can give two variations different from conjugation for the same reason (Dahdah, 1992).

The AVM model for verbs is illustrated in Figure 8.

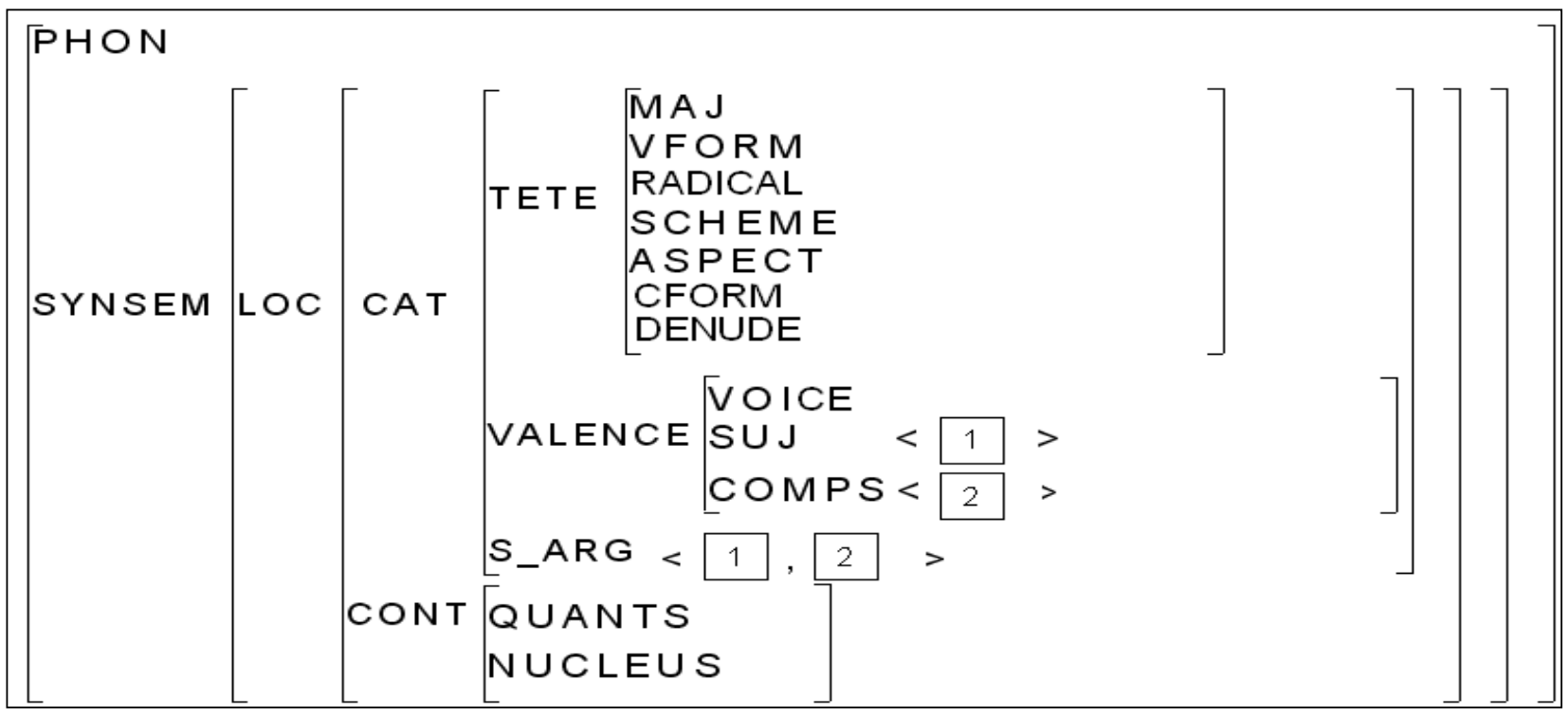

**Figure 8:** Model of an AVM of a verb

The morphological verb features are always given in the feature HEAD, the syntactical ones in the feature VALENCE and the semantic ones in the feature CONT. In the following figure we give an example of a verb AVM using *kataba*.

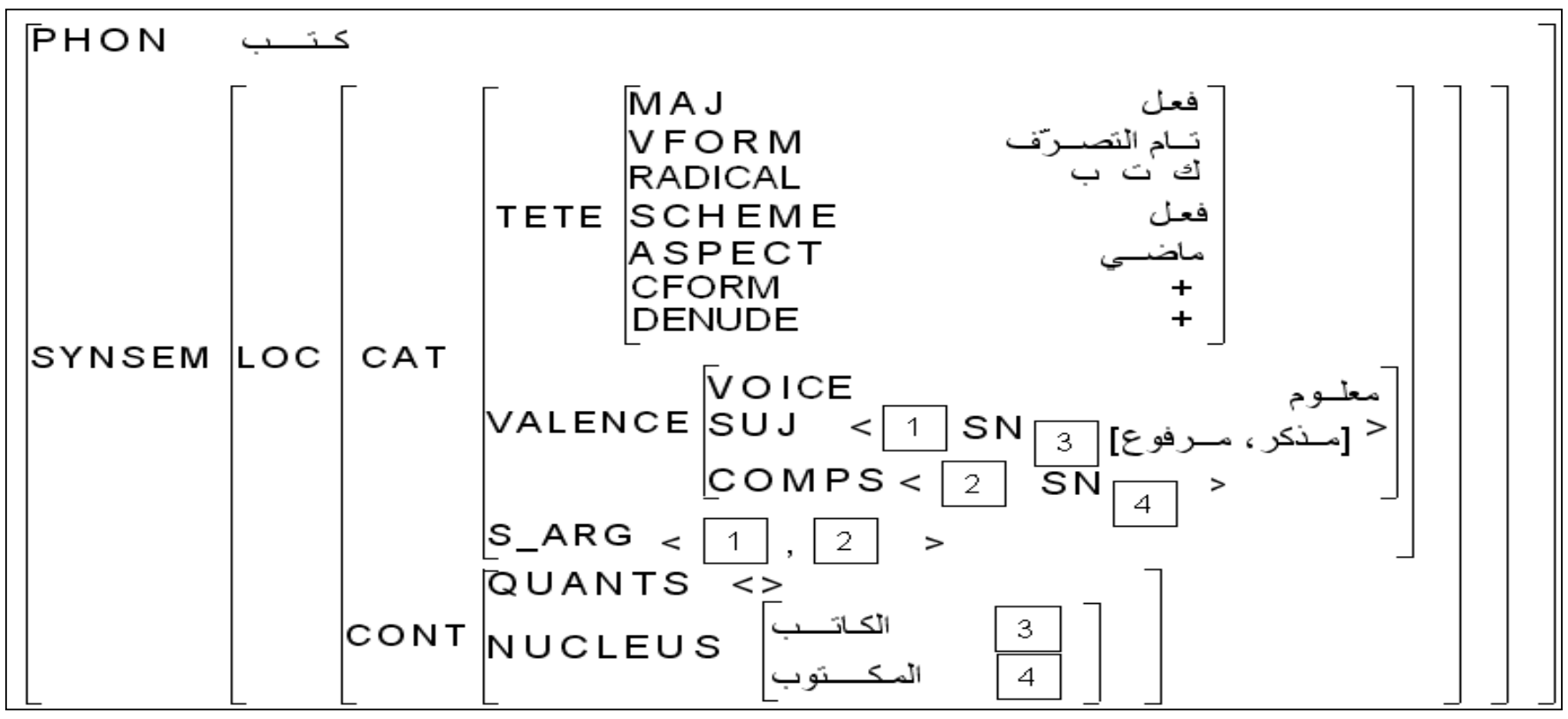

**Figure 9:** AVM of the verb *kataba*

The example in Figure 9 shows that the verb *kataba* is conjugated in the perfect tense, in active voice and has as a root *ktb*. This verb can subcategorise a subject and an object. These values are contained in the feature S-ARG describing a structure list. This feature is considered as a valence feature concatenation. In addition, we remark that a subject carries a specification on its index: the nominal category should be masculine and nominative.

[3] the verb appears in its canonical representation (as opposed to "augmented")

## 4 LMF MODEL

After presenting the HPSG formalism adapted for the Arabic language and defining the appropriate AVM for every lexical category (noun, particle and verb), we now describe the ISO LMF specification platform under the specific perspective of the projection of lexical structures. Through this study we can understand LMF specificities and subsequently identify the common points that are processed by the two abstract models (HPSG and LMF). As a result we can extrapolate a method allowing the projection from HPSG lexicons into LMF.

The objective of LMF is to propose a modular data model that is independent from any particular lexicographic theory and allows abstraction from concrete representation (e.g. proprietary syntax, XML structure based on the TEI guidelines, database model, etc.). The modelling framework, initially experimented with in the terminological domain (Romary, 2001), operates at the conceptual level: it aims to identify the essential components of a generic lexicographic model, to describe the constraints governing their arrangement, and to identify the descriptors (data categories) that are associated with them. The LMF standard is based on a core model together with a set of five extensions, as explained in the following subsections.

### 4.1 Core part

The core model of LMF specifies the concepts of lexicon, word, form and sense in keeping with a semasiological view of lexical structures[4]. It describes information concerning a lexicon and the basic hierarchy of the information that can be included in a lexical entry. The core model is illustrated in Figure 10.

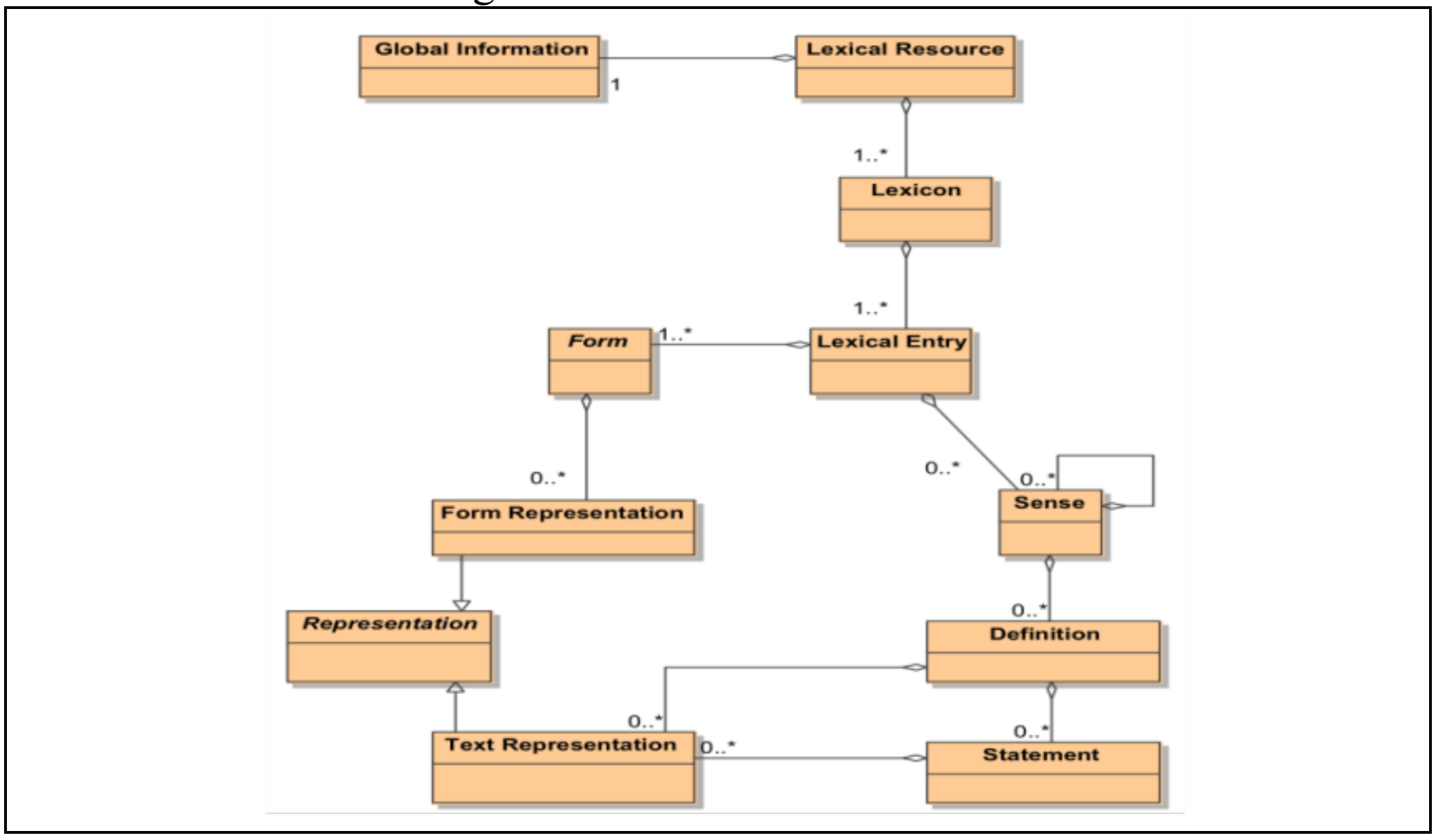

**Figure 10:** LMF core model

[4] Similarly, the TMF standard (ISO 16642) is dedicated to onomasiological structures as encountered in conceptual systems and terminologies.

In Figure 10, the Lexical Resource component is a singleton that represents the entire resource, seen as a container for one or more lexicons. The Lexicon component is the informational locus for all lexical entries of a source language within the database. A lexicon must contain at least one lexical entry and must not allow certain subclasses. The Global Information component contains the administrative information and other general attributes of a lexicon (e.g. the metadata associated to a lexical resource). The Lexical Entry component may represent a word, a composed expression, or an affix in a given language.

With the semasiological perspective in mind, the Lexical Entry component instantiates the link between the Form and Sense components. A lexical entry may have one or several different forms and may have none or several different meanings. The Entry Relation component allows one to represent cross-references between two or more lexical entries within or across lexicons. It can contain attributes that describe the type of relationship.

The LMF core model can be extended to satisfy further requirements bound to the treatment of specific lexicographic aspects. Several possible extensions are described in the LMF standards, among which we may mention the morphological extension, the syntactic extension, the semantic extension, the inflectional paradigm extension and the multilingual annotations extension. These extensions must be selected according to the needs of the designer of a specific lexical model. In our case we will put a specific emphasis on the morphological, syntactic and semantic extensions, as presented in the following sections.

## 4.2 Morphological extension

The goal of this extension is to provide mechanisms that support the development of the NLP lexicons describing the morphology of the lexical entries.

**Example 4.1 (The Arabic word *'ayn* [eye]):**
The object diagrams of Figure 11 and Figure 12 represent two different ways to describe the inflectional part of the Arabic word ‘*ayn* (eye).

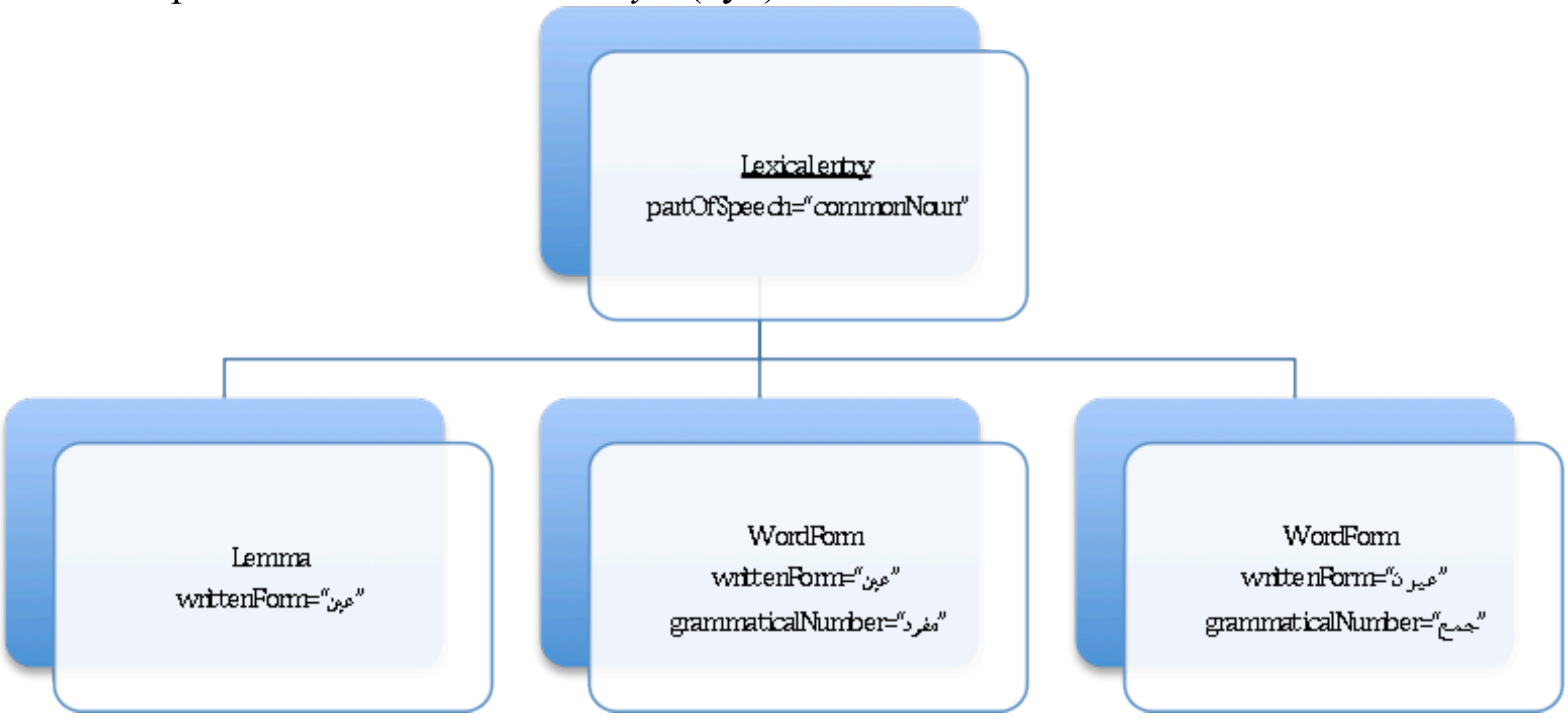

**Figure 11:** Objects diagram representing the inflectional part of *'ayn* "ع-ي-ن" without inflectional paradigm

As mentioned earlier, the LMF structure depicted in Figure 11 can be implemented in any specific format and in particular may be serialised according to any kind of XML representation as long as it is isomorphic to the underlying LMF model. In the rest of the paper, we will more specifically apply our examples using the Text Encoding Initiative (TEI) framework, benefiting from a widely accepted background for our concrete representation, and also making full use of the customisation facilities offered by the TEI infrastructure. The elementary lexical structure presented in Figure 11 can easily be serialised in TEI, as follows[5]:

```
<entry>
    <gramGrp>
      <pos>commonNoun</pos>
    </gramGrp>
    <form type="lemma">
      <orth>عين</orth>
    </form>
    <form type="inflected">
      <orth>عين</orth>
      <gramGrp>
        <number>مفرد</number>
      </gramGrp>
    </form>
    <form type="inflected">
      <orth>عيون</orth>
      <gramGrp>
        <number>جمع</number>
      </gramGrp>
    </form>
  </entry>
```

Note that in Figure 11 above, two inflected forms of the singular word *‘ayn* (eye) and of the plural word ‘*ayūn* are represented without passing through an inflectional paradigm. In this case, every inflected form must be described in an object of the class InflectedForm.

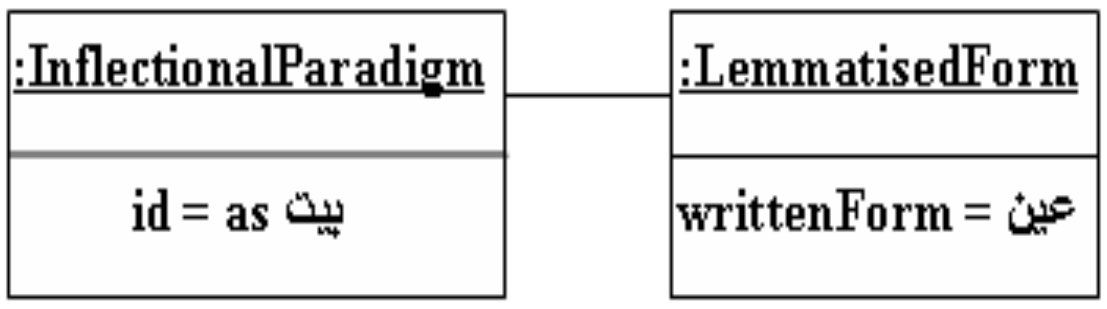

**Figure 12:** Diagram of objects representing the inflectional part of *'ayn* with inflectional paradigm

In Figure 12, the two inflected forms of *‘ayn* must be generated automatically using the inflectional paradigm. The paradigm used called "*as bayt*" (house) consists of inserting the letter *ū* in the fourth position of the word *‘ayn*. It can be shared with other lexical entries where the inflectional part is like *‘ayn* (e.g. *bayt* and *bayūt*).

## 4.3 Syntactic extension

The syntactic extension of the LMF standard aims at providing ways to describe the word properties when combined with other words and phrases in a sentence.

**Example 4.2 (the Arabic word *kataba*):**

[5] TEI elements belong to the namespace http://www.tei-c.org/ns/1.0

The verb *kataba* (wrote) subcategorises a subject that must be a nominal phrase (NP) and an object that must also be a nominal phrase. This syntactic behaviour is described in Figure 13, taking into account that a verb can admit more than one syntactic behaviour.

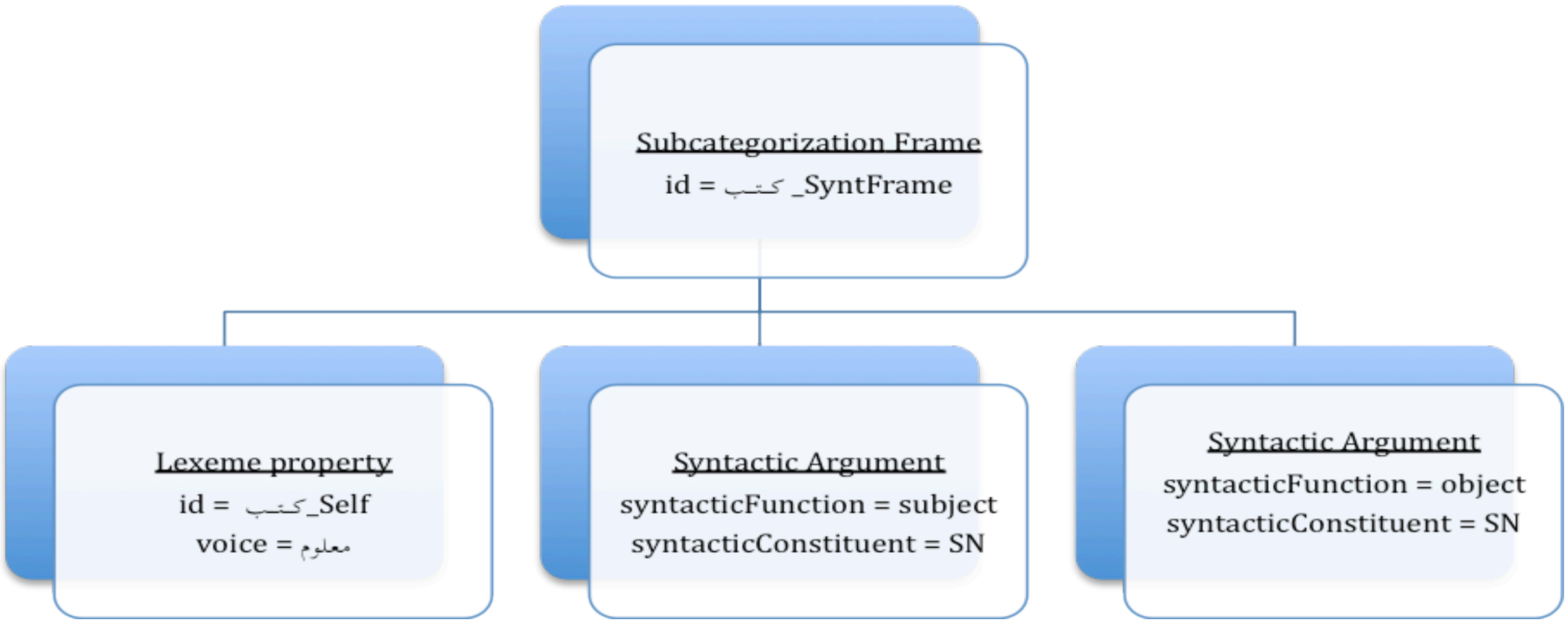

**Figure 13:** Diagram of objects representing the syntactic behaviour of the verb *kataba* "كتب".

Figure 13 shows how the syntactic behaviour is represented in an object of the class Sub-categorisation Frame. When a verb has more than one frame, each version of this verb is considered as a new entry and will be projected in LMF differently (*kataba alwaladu* and *kataba alwaladu risālata*). This object is combined with as many objects of the class Syntactic Argument as the number of constituents of the verb *kataba* requires.

## 4.4 Semantic extension

With the semantic extension of LMF, it is possible to describe a semantic profile together with the relations with other meaning within the lexical database. The extension also provides the means of linking syntactic and semantic description, typically at the argument level.

**Example 4.3 (the Arabic word *kataba*):**

In the example described below, we present an object diagram illustrating the relationship between the syntactic and the semantic part of the verb *kataba* (wrote).

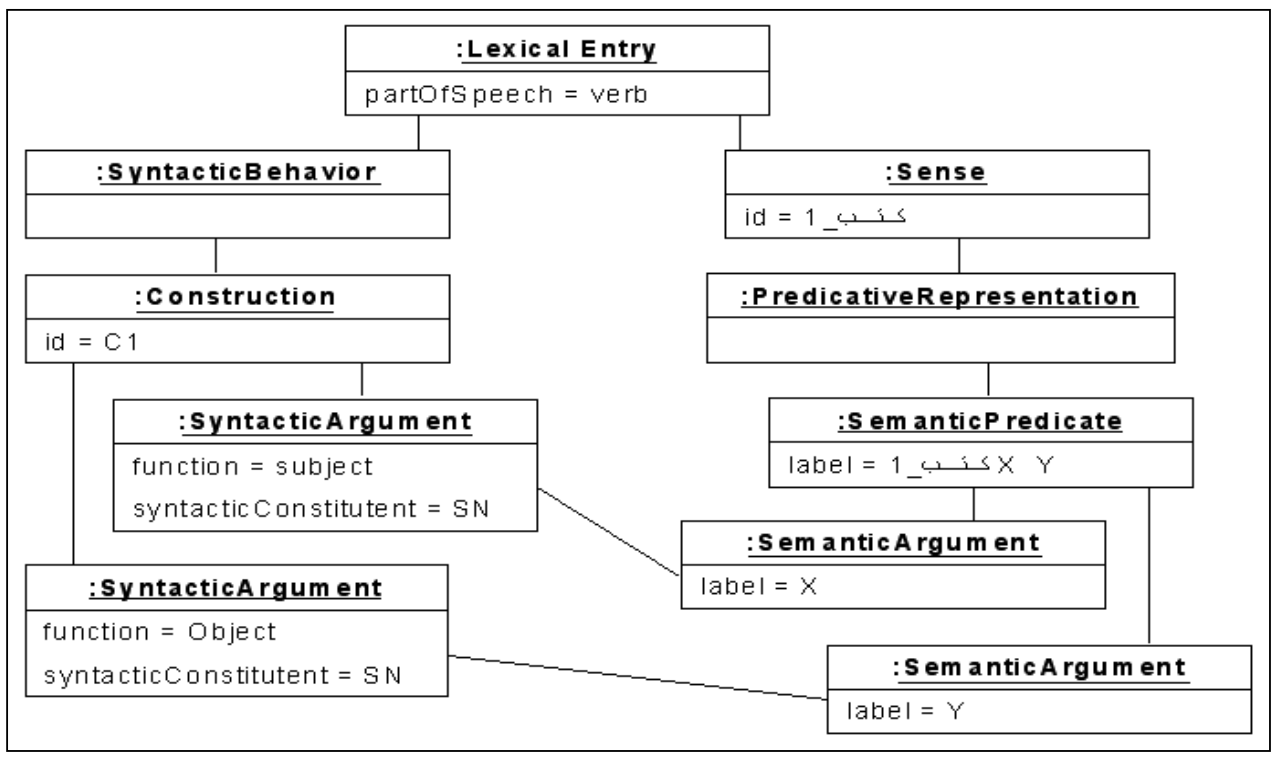

**Figure 14:** Object diagram representing the relationship between the syntax and the semantic of the verb *kataba*

In Figure 14, the subject is labelled as X and the object as Y. If we suppose that X represents *al-walad* (the boy) and Y *al-dars* (the lesson), then the association of these constituents with the verb *kataba* gives the significance of *the boy wrote the lesson* expressed in the object SemanticPredicate.

From the overview of the basic LMF mechanisms presented above we can see how sub-categorisation phenomena, which are essential in the HPSG formalism, can be taken into account in the LMF standard. The main difference between the two representation models essentially resides in the manner in which the lexical entries are actually organised. A canonical (or derived) form with all its inflected forms constitutes one single lexical entry in LMF. In HPSG, however, each form, whether it is derived, canonical or inflected, constitutes a unique lexical entry. We can also identify features in HPSG that are specific to the Arabic language and have no equivalent in LMF as it stands as a published standard. For these, we will have to provide specific extensions by describing new data categories, which will then be submitted to the Data Category Registry (ISOCat.org). For instance, most data categories presented in section 3 for the morphological description of the Arabic language have at present no equivalent in ISOCat.

## 5 PROJECTING HPSG LEXICAL STRUCTURES IN LMF

In this section, we present the proposed method for the projection of a syntactic HPSG lexicon into an LMF compliant representation. This method is designed on the basis of the LMF meta-model and on the above-mentioned extensions applied to this model, incorporating the characteristics of the HPSG theoretical framework. The method that we propose is articulated around two essential steps, namely the identification of a projection rule system and the projection process itself.

### 5.1 Identification of projection rule system

The first phase consists of studying the various lexical categories represented in HPSG in order to identify the nature and the information associated to each feature of an AVM adapted to the Arabic language. During projection, each such feature will be transformed into an LMF class attribute. The intrinsic nature of a feature — whether morphological, syntactic or semantic — helps us to know to which LMF component the feature is going to be projected. We can then limit the number of the classes that will be affected by the projection accordingly.

The feature type (e.g. morphological, syntactical) helps us to identify in which class the projection is going to be made. If we take the case of the feature RADICAL, the feature keeps the same value for the canonical (or derived) form and its inflected forms. We can say, therefore, that it is a feature that relates to the class LexicalEntry. However, if we take the case of the feature SCHEME, we note that this feature changes from an inflected form to another and in this case it relates to the class InflectedForm. In the next paragraph, we present the specific rules that we have identified for the morphological features.

#### *5.1.1 Projection rules for morphological features*

The rules corresponding to morphological features are divided into two types: those that can be applied to all lexical categories (noun and verb, particle, non-inflected noun and non-inflected verb) and those that may only be applied to specific categories.

<u>*Example of a rule applicable to verbs only:*</u>

| |
|---|
| *$R_{1m}$ : (Feature$_{HPSG}$=PHON) ∧ (Value(SCHEME)∈FDC)→in LemmatisedForm: att=lemma ∧ val=Valeur(PHON) ∧ in InflectedForm: att=orthography ∧ val=Value(PHON)* |

In the rule $R_{1m}$, FDC designates the set of all models representing the canonical and derived forms relative to a verb (FDC ={CaCaCa, CaCCaCa, CaCaCaCa,CaaCaCa,CaCaCaCaCa, CaCaaCaCa,CiCCaCaCa, CiCCaCaCa, CaCCaCaCa, CiCCaCCaCa, CiCCaCCaCa, CiCCaaCaCa, CaCaCCaCa, CiCCaCaCa, CiCCaCaCaCa} { فعل، فعلل، فعّل، فاعل، أفعل، تفعّل، تفاعل،انفعل، افتعل، افعلّ، استفعل، افعوعل، افعولّ، افعالّ، تفعلل، افعللّ، افعنلل}). Note that the function Value allows returning an HPSG feature value. We can take the case of the verb *'akhraja* أخـرج (to extract). The model for this verb is *'af'ala* أفــعــل and belongs to FDC. Therefore, after having applied the rule $R_{1m}$, a new attribute is added in the class LemmatisedForm and named lemma and the value is equal to *'akhraja* أخـرج, and another is added in the class InflectedForm with the name orthography, and its value is equal to *'akhraja* أخـرج.

<u>*Example of a rule applied only to nouns:*</u>

| |
|---|
| *$R_{3m}$: (Feature$_{HPSG}$=PHON) ∧ (Value(NOMB)=SINGULAR) ∧ (Value(GENR) = MASCULIN) ∧ (Value(DIMINUTIVE)=non diminutive) → in LemmatisedForm: att=lemma ∧ val=Value(PHON) ∧ in InflectedForm: att=orthography ∧ val=Value(PHON)* |

The rule $R_{3m}$ is applied to the canonical or derived forms of a noun. The application of this rule results in the addition of two new attributes: the first one is added to the class LemmatisedForm and named lemma and has as its value the HPSG feature PHON, and the second is added to the class InflectedForm. The second attribute is named orthography and has as its value the HPSG feature value.

<u>*Example of a rule applicable to verbs and nouns only:*</u>

| |
|---|
| *$R_{5m}$: ∀ Feature$_{HPSG}$ ·∃ attribute$_{LMF}$: attribute$_{LMF}$ ≡ Feature$_{HPSG}$ ∧ ¬ Variable(Value (Feature$_{HPSG}$)) → in LexicalEntry: att = attribute$_{LMF}$ ∧ val = Value(Feature$_{HPSG}$)* |

The rule $R_{5m}$ is applied to the features that always take the same values for the canonical form (or derived) and its inflected forms. Let us note here that the function Variable is a function that returns true if a feature keeps the same value for the canonical form (or derived) and all its inflected forms. Figure 15 represents an example of the application of the rule $R_{5m}$.

| Canonical form | Value(MAJ) |
|---|---|
| ذهب | verb |

| Inflected form | Value(MAJ) |
|---|---|
| ذهب | verb |
| ذهبت | verb |
| ذهبا | verb |
| ذهبنا | verb |
| ذهبوا | verb |
| ... | verb |

**⟹ Variable (Value (MAJ))=false ⟹ $R_{5m}$**

**Figure 15:** Example of the application of the rule $R_{5m}$

We can observe that in Figure 15, the value of the feature MAJ remains unchanged for the verb *dhahaba* ذهب (to go) and for all its inflected forms. In this case, we must apply the rule $R_{5m}$ since the feature MAJ has its equivalent in LMF that is equal to the attribute PartOfSpeech.

*Example of a rule applicable to particles, non-inflected nouns and non-inflected verbs:*

| |
|---|
| $R_{9m}$: *$Feature_{HPSG}$ = MAJ → in LexicalEntry: att = GrammaticalCategory ∧ val = Value(MAJ)* |

The rule $R_{9m}$ is applied only to the features MAJ and PHON given that these features exist in any type of particles.

### 5.1.2 Identified rules for syntactic features

The identified rules for syntactic features are considered to be paradigms. Several lexical entries can have the same syntactic behaviour and in this case they share the same projection rule through their identifier. Rule $R_{1syn}$ is an example of this in a case where the value of the HPSG feature can have more than one value at a time (complex). This rule is defined formally as follows:

| |
|---|
| *$R_{1syn}$: Complex Value($Feature_{HPSG}$) → in SyntacticArgument: att = function ∧ val = function($Feature_{HPSG}$) ∧ att = SyntacticConstituent ∧ val = Value($Feature_{HPSG}$)* |

Among the features to which we apply the rule $R_{1syn}$ are SPR, TOPIC, ATTRIBUT and COMPS. Note that rule $R_{1syn}$ must be applied as many times as there are values for the feature in question.

The features SUJ and COMPS will be projected by using rule $R_{1syn}$ because their values are composed. On the other hand, VOICE will be projected by using rule $R_{2syn}$ as this feature admits its equivalent in LMF and its value is simple:

| |
|---|
| *R2syn : atomic (value (attributeHPSG)) ∧ ∃ attributeLMF : attributeLMF≡ attributeHPSG ® in : Self att = name (attributeLMF) ∧ val = value (AttributeHPSG)* |

### 5.1.3 Projection rules for semantic features

Semantic features are represented in the feature CONT, which contains a list of quantifiers. The semantic part, which we consider here, is represented by the feature NUCLEUS whose value is generally an AVM composed of the features agent-noun and patient-noun if it is about a verb, but is empty otherwise. So far we have identified only one projection rule applicable to the semantic features illustrated by $R_{1sem}$.

| |
|---|
| *$R_{1sem}$: if Nucleus ≠ <> → in SemanticArgument: att=agent-noun ∧ val=value(agent-noun)* |

The same rule $R_{1sem}$ can be applied to the feature patient-noun. The attribute will be projected to the class SemanticArgument. Note that a lexical entry projection must be made in the appropriate locus (i.e. component) of the LMF model. The lexicon under work already contains other lexical entries that have been projected. If we take the case of the verb *dhahaba* (he goes) and the inflected form *dhahabnā* (we go), we observe that in HPSG these two lexical entries have two independent AVMs. Whereas, at the time of the projection, the two entries only represent one lexical entry of which *dhahaba* (he goes) is a canonical form and *dhahabnā* (we go) its inflected form.

### 5.2 Projection process

The projection phase, the goal of which is to apply the corresponding projection rules to all features characterising a lexical entry, is based upon three essential stages. These stages are applied iteratively on all lexical entries included in the lexicon. These stages are illustrated in Figure 16.

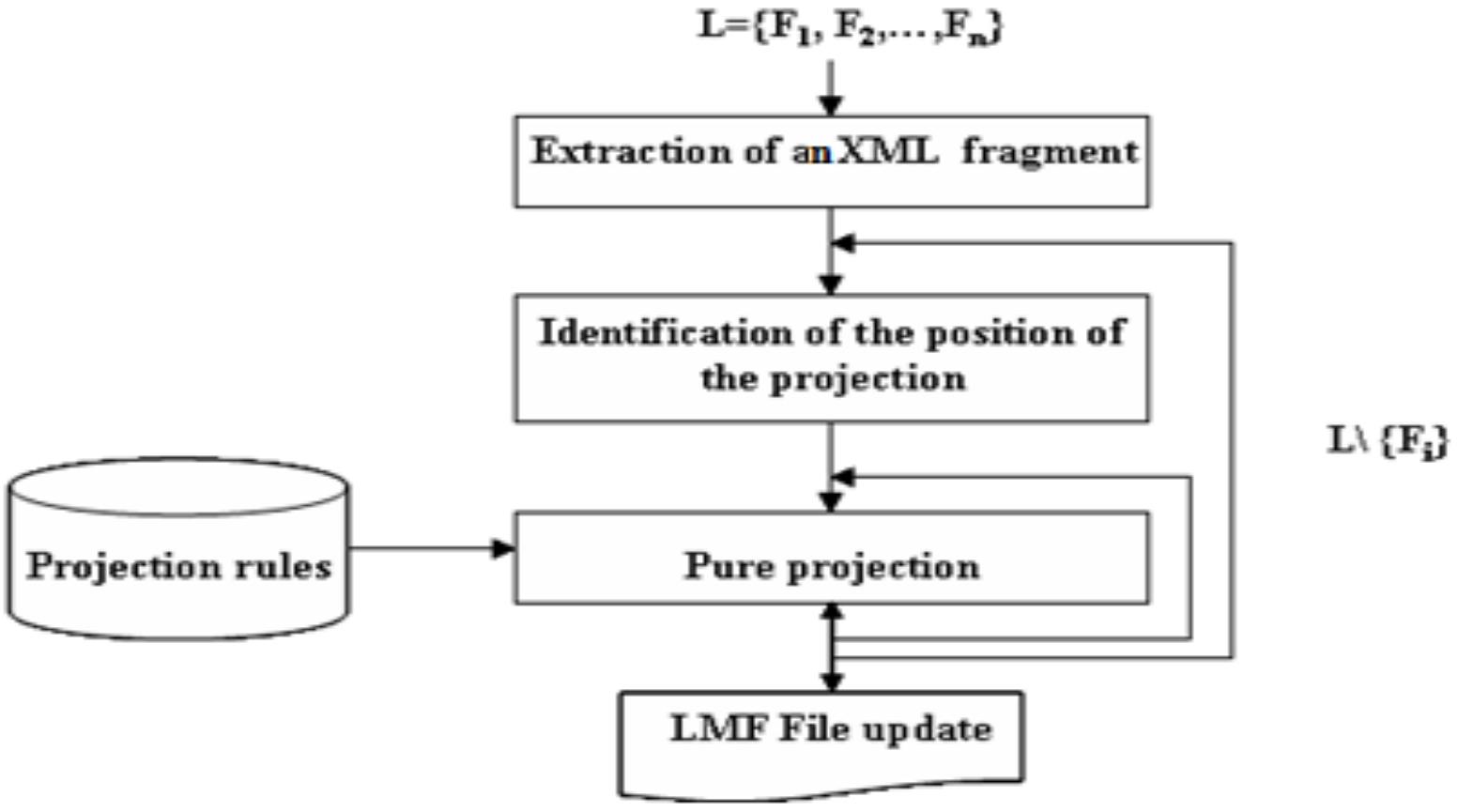

**Figure 16:** Stages of proposed method

The input for our process is a set of lexical entries that can represent verbs, nouns, particles or a combination of these categories. A projection starts with the first open lexicon. In the following paragraphs we are going to give an idea of the method stages required for the extraction of every lexical entry, its XML fragment, the identification of its projection position and the projection using the adequate rules.

#### *5.2.1 Extraction of XML fragments for AVMs associated with lexical entries*

The first phase consists in extracting the XML file fragment, which represents the lexical entry AVM to be projected. A fragment extraction phase is essential because the projection is made on a word by word basis. Figure 17 illustrates this stage.

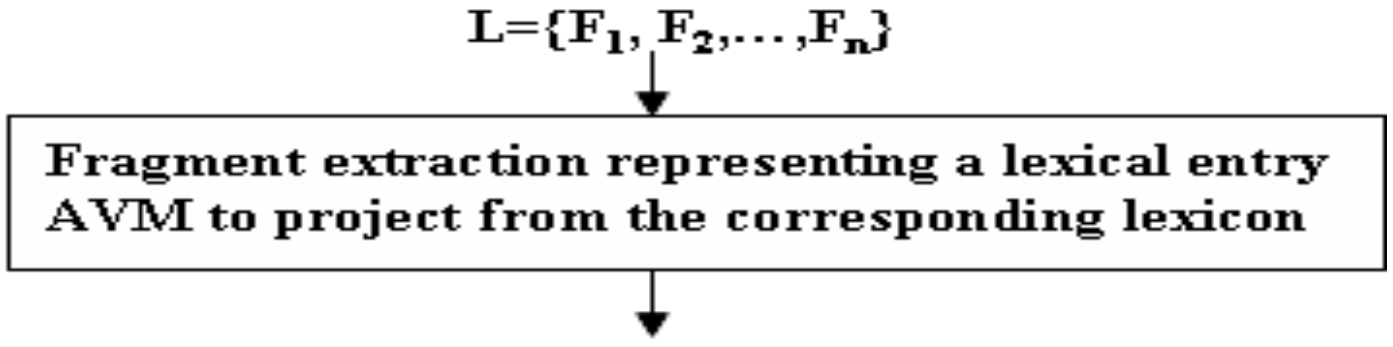

```xml
  <?xml version="1.0" encoding="UTF-8" ?>
- <Lexique>
  - <fs>
    - <f name="PHON">
        <string>اخرج</string>
      </f>
    - <f name="SYNSEM">
      - <fs>
        - <f name="LOC">
          - <fs>
            - <f name="CAT">
              - <fs>
                - <f name="TETE">
                  - <fs>
                    - <f name="MAJ">
                        <symbol value="verbe" />
                      </f>
                    - <f name="VFORM">
                        <symbol value="تام التصرف" />
                      </f>
                    - <f name="RADICAL">
                        <symbol value="خ ر ج" />
                      </f>
                    - <f name="SCHEME">
                        <symbol value="افعل" />
                          .
                          .
                          .
    </f>
  </fs>
```

**Figure 17:** XML fragment extraction of a lexical entry in conformity with ISO 24610-1

The example in Figure 17 concerns the verb *akhraja* أخرج (to take out). At this stage the description of the various features characterising this verb is encoded according to the ISO-TEI standard for feature-structures (ISO 24610-1).

### *5.2.2 Identification of projection position*

The projection basic algorithm uses some tests that concern the verification of the lexical entry form to be projected and the position of the projection. Figure 18 illustrates the position of these tests at the time of the projection process.

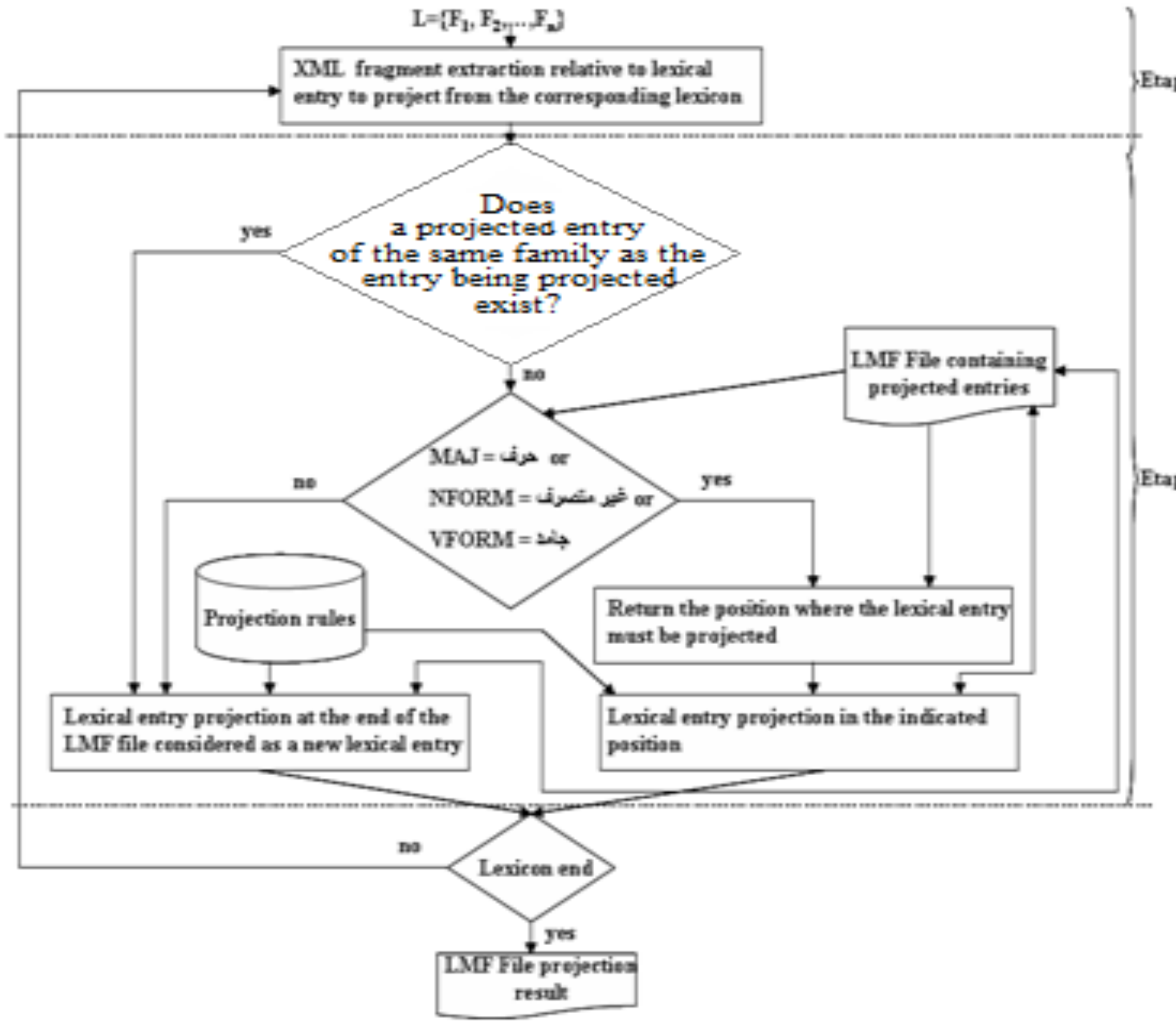

**Figure 18:** Overview flow chart

As Figure 18 indicates, the process first consists of verifying if the lexical entry under projection can admit inflected forms or not. This verification is essential as in the case where inflected forms exist it is necessary to know in which position the projection must be made. We need to remember that in LMF a lexical entry is composed of the canonical or derived form and all its inflected forms whereas our starting point is a lexicon containing different lexical entries that can be canonical, derived or inflected forms represented according to the HPSG formalism. These entries are organised according to the choices made by the lexicon's designer. Therefore, the LMF compliant output contains a number of lexical entries that must be lower or equal to the number of existing lexical entries in the HPSG lexicon. Let us note that features that allow us to know if a lexical entry admits the inflected forms or not are NFORM and VFORM for nouns and verbs respectively. For particles we have no inflected forms.

The process then moves to projection position verification. If the lexical entry to be projected can admit inflected forms, it is necessary to browse the LMF file containing the projected lexical entries to know if a lexical entry of the same class has been projected. This research is based on:

- the values of the features RADICAL and DENUDE in the case of a verb representing a canonical form or an inflected form of a canonical form,
- the values of the features RADICAL and SCHEME for the rest of the verbs,
- the values of the features NATURE and RADICAL for the nouns.

### 5.2.3 *Pure projection*

The phase "pure projection" consists of browsing the XML file already extracted in the previous phase in order to extract features representing the lexical entry to project. On the basis of the survey of the various AVMs already done in the first stage, we now know the different features forming every lexical category and can apply the corresponding projection rule. Figure 19 illustrates this stage.

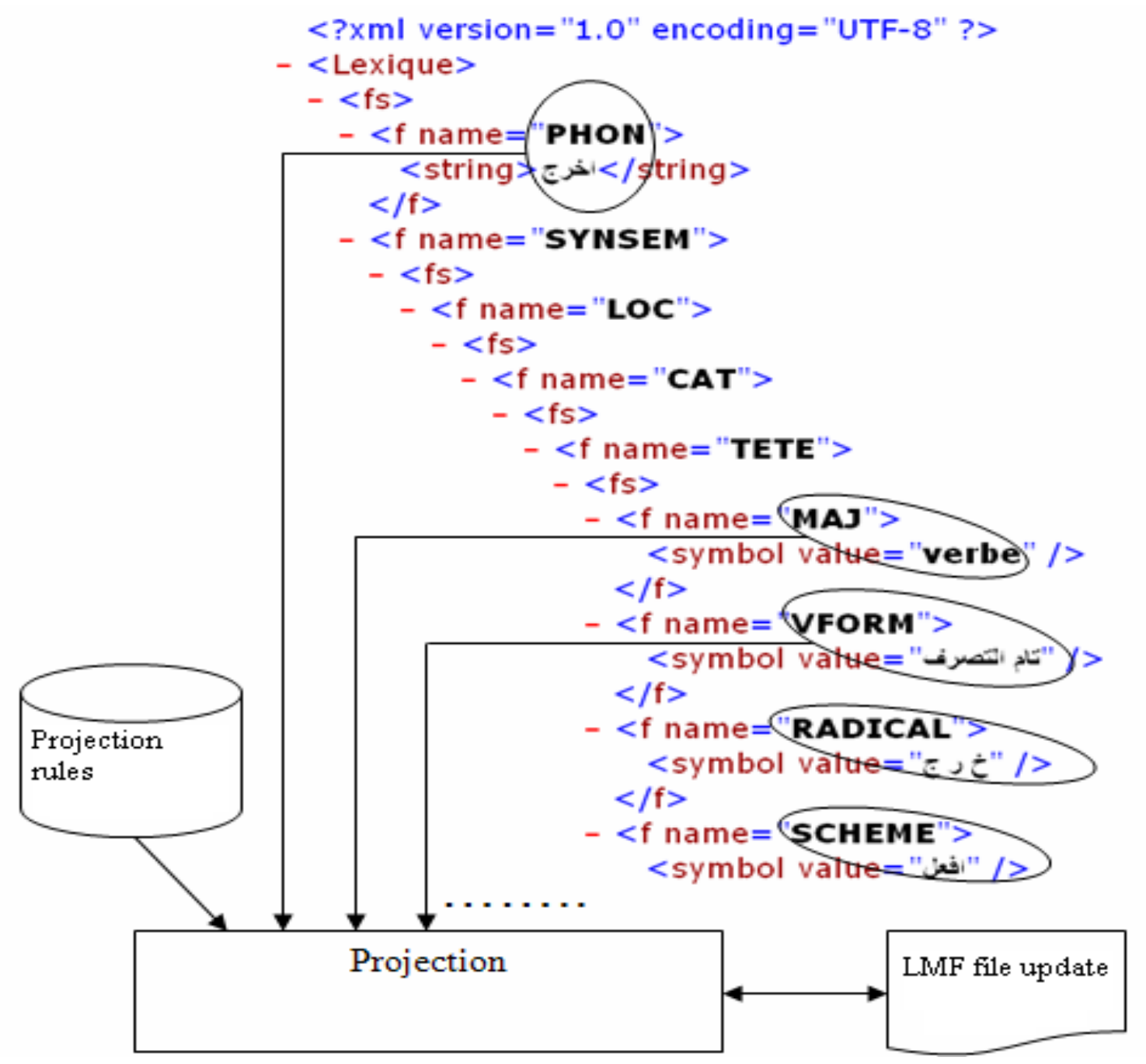

**Figure 19:** Projection of a word's features AVM

For every lexicon entry, we extract a feature together with its value and project them using the adequate projection rule until arriving at the end. The example in Figure 19 relates to the verb *akhraja* "أخـرج" (to take out).

The proposed method is independent from the HPSG lexicon organisation. The order of the lexical entries in the lexicon does not have any impact on the projection. We can find, for example, in the HPSG lexicon a canonical form before its inflected forms or the opposite. Also, the addition or the adoption of an HPSG feature does not influence the result obtained from the projection. Our established projection rule system processes all possible cases that can arise in the Arabic language. The projection of another HPSG lexicon using another language than Arabic is possible. It is sufficient to modify some projection rules in accordance with the particularities of the new language.

This method helps us to then process both the conception and the implementation phases, which is the subject of the following section.

## 6 THE IMPLEMENTATION PHASE

Having presented the proposed method for the projection of HPSG into LMF in the previous section, in this section we are going to describe the achieved prototype in order to validate this method.

### 6.1 General architecture of the achieved prototype

The prototype allows projection of one or several existing HPSG syntactic lexicons into LMF. The projection will give us a normalised representation of these lexica and therefore encourages their merging. Our prototype is composed of two modules. The first concerns the projection phase and is applied after having chosen and opened one or several HPSG lexicons. The second concerns the generation of the LMF file resulting from the projection. Figure 20 depicts these different modules.

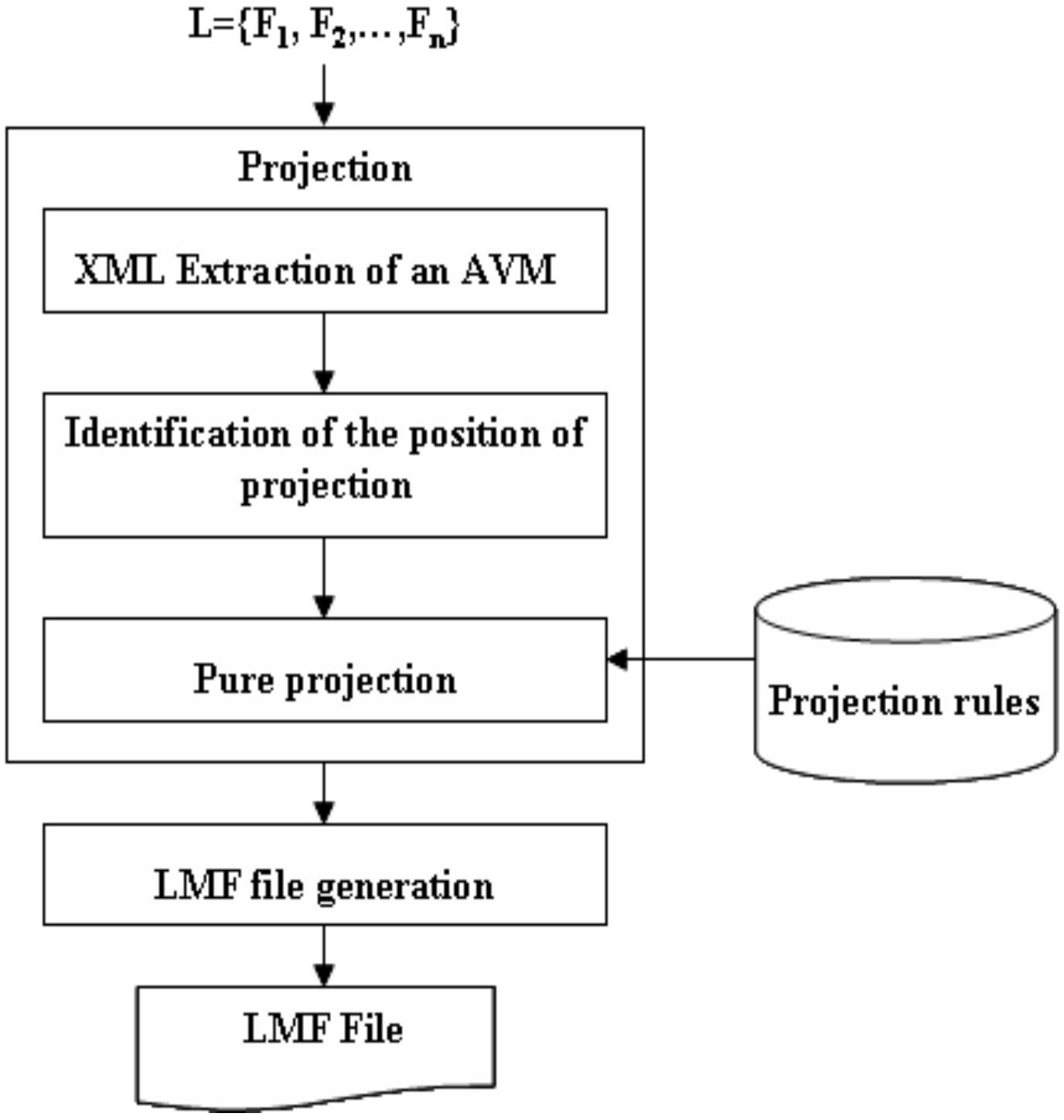

**Figure 20:** Prototype's architecture

In order to execute the projection, the user must open at least one lexicon that is represented in HPSG. The system will then browse every open lexicon entry by entry to extract the XML fragment relative to the corresponding entry. For every extracted XML fragment the system also extracts every attribute and its value and projects them using the base of the suitable projection rules.

The HPSG lexicon to be projected is in turn composed of one or several AVMs. An AVM is itself composed of features and values. A feature value can be a simple value or a composed value (list or AVM). As for the LMF lexicon, the result of the projection is composed of a set of elements. Every element can be composed of other elements and/or the data categories (DC). Every data category constitutes of an attribute having a value.

### 6.2 Some prototype functionalities

The implemented prototype allows the projection of a lexicon represented in XML and respects the standard format of representation of feature structures introduced (Hasni *et al.*, 2006). Figure 21 illustrates the projection process.

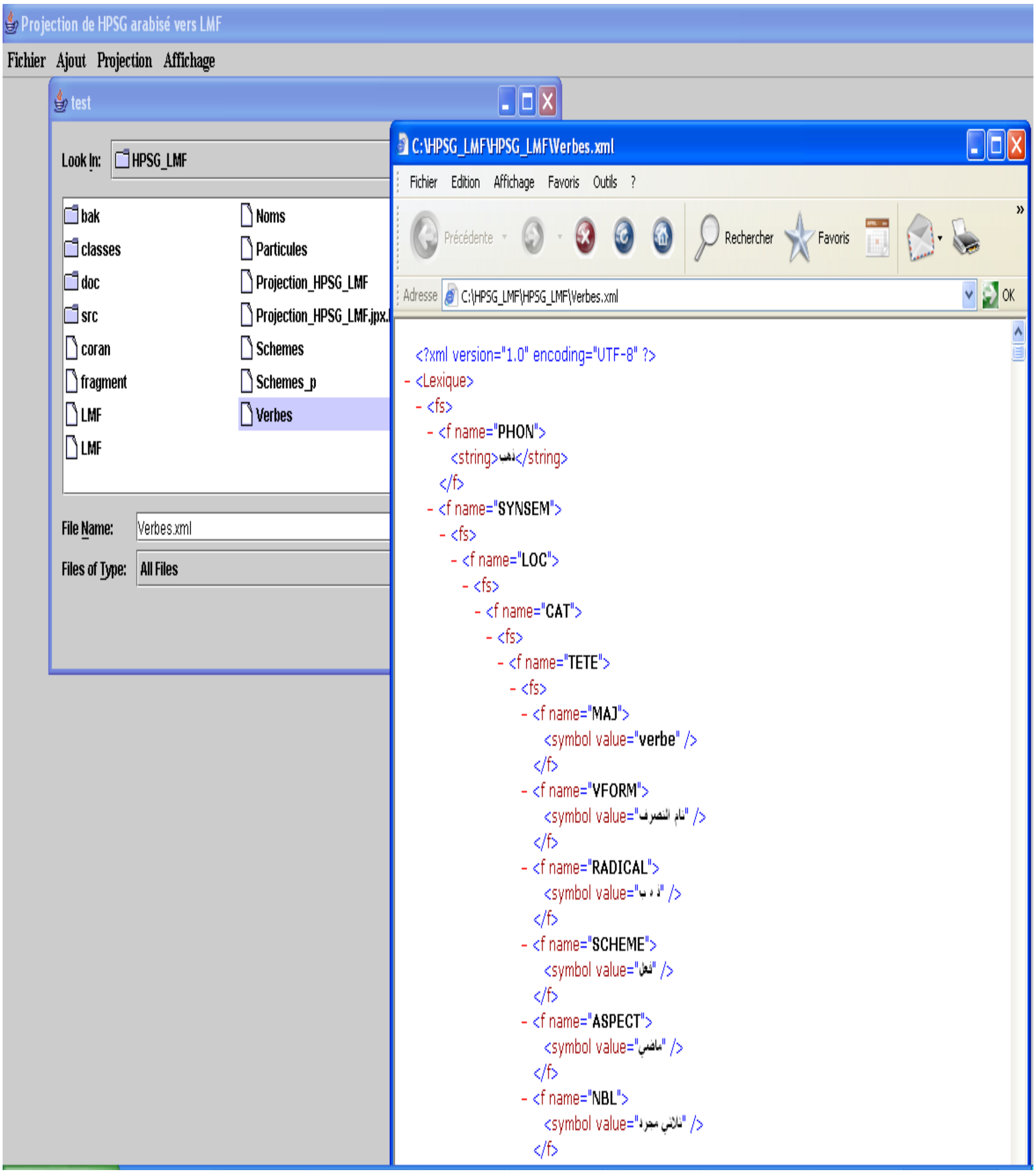

**Figure 21:** Opening of the file "Verbes.xml"

The displayed file in Figure 21 corresponds to the projection result file. It reflects the application of the rules relative to a verb that we have detailed previously. This file takes into account the DTD that is represented in ISO/TC37/SC4 N130 rev.9 2006.

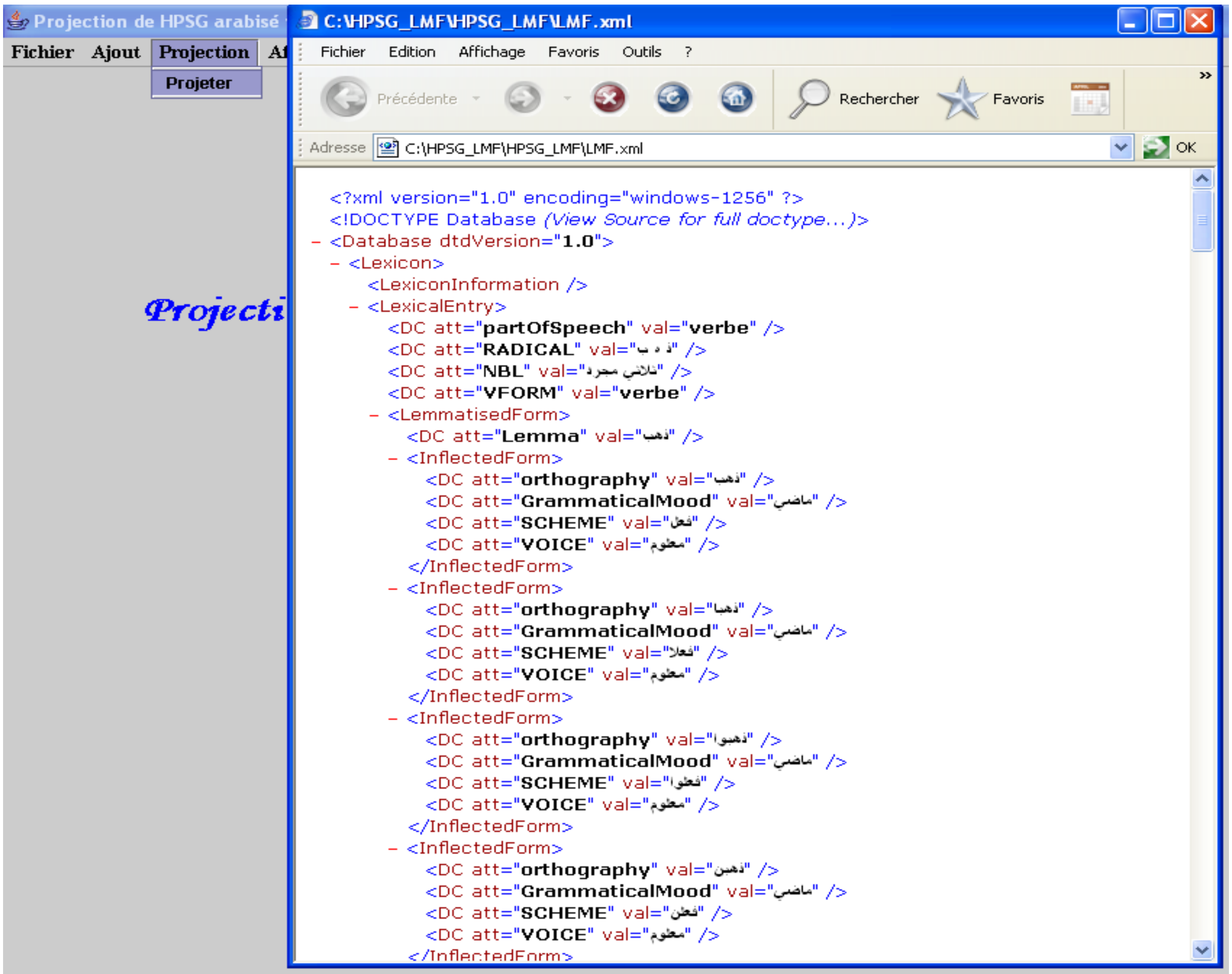

**Figure 22:** Projection of the file “Verbes.xml”

The menu in Figure 22 displays the lexical entries of the constructed files. Figure 23 is an example of the verbs that are found in the file "Verbes.xml".

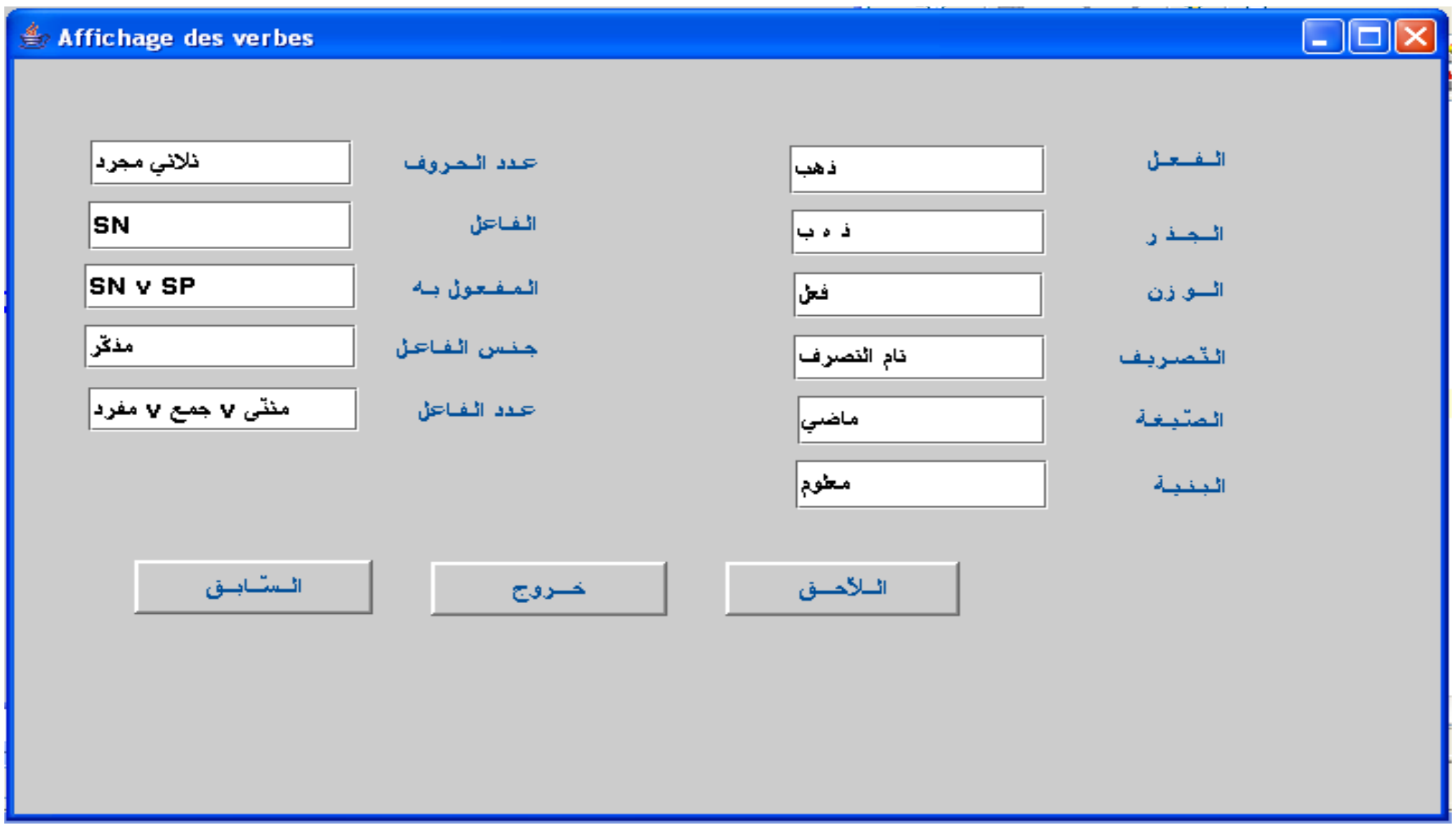

**Figure 23:** Display of the entries contained in the file "Verbes.xml"

The graphical interface of Figure 23 shows the feature values of a verb from the lexicon in question. This interface also provides navigation facilities in the lexicon by showing the characteristics of the verbs that precede and follow the verb shown.

### 6.3 Evaluation

In order to evaluate the constructed system we projected 10 Arabic HPSG lexicons into LMF language. Projected lexicons have varied structures and contents and allow the obtaining of a normalised lexicon in conformity with LMF and without any loss of information. These lexicons contain different features that we have added to Arabic adapted HPSG in order to bind every canonical or derived form to its inflected forms. These lexicons can also contain canonical and/or derived forms without inflected forms. The obtained LMF lexicon contains 3,000 verbs, 450 nouns and 50 particles.

A Lexicon projection result in conformity with LMF can result in loss of information because projected lexicons possess features that do not exist in the base of chosen features. Data categories in HPSG are not standardised and every user can define his proper data in order to achieve his goal. Therefore, the same feature can exist in several HPSG lexicons under different writing formats. HPSG lexicons that generate some lexicons that do not conform to LMF and result in loss of information are those that contain inflected forms and do not use the features that we have already added.

We can deduce that to get a lexicon in conformity with LMF without any loss of information necessitates three conditions in the HPSG lexicon source of projection: the first of these is to add the features that bind canonical or derived forms to their inflected forms in the case where the HPSG lexicon to be projected contains inflected forms; the second is to add all HPSG features in the feature basis; the third is to reject all schemes of triliteral verbs in order to avoid conflict between two verbs that may be written in the same way *kharaja* (خَرَجَ) and *kharija* (خَرِجَ).

Our system may also be considered to be extensible. We opted for a simple design assuring module autonomy and we have implemented a projection prototype of the HPSG into LMF in an object oriented language encouraging the use of expandable software. We can thus extend our work by the addition of projection rules that allow the use of lexicons belonging to other formalisms. However, prototype portability is not assured as it is designed to manipulate the Arabic language and to use only Windows operating systems that support this language. Our achieved prototype permits not only the recuperation and the fusion of HPSG lexicons without data redundancy but also allows processing of several variations such as the orthographic variation. At the lemma level, for two etymologically bound forms having identical pronunciation and belonging to different inflectional paradigms, our system contains two distinct and separate lexical entries. Furthermore, the prototype allows projection of lexical entries that are categorised as grammatical words. We find in the noun category pronouns (e.g. personal, demonstrative), proper nouns, abstract nouns, etc. in the particle category, we find the significance letters (e.g. conjunctions) and the construction letters.

## 7 CONCLUSION AND PERSPECTIVES

In this article we have developed a system allowing the projection of an HPSG syntactic lexicon into an LMF compliant lexical model. This system allows us to project lexical entries of different lexical categories from any HPSG lexicon. This projection will help us to either recover some existing HPSG lexicons, or to merge them and/or to integrate them with other lexicons in order to create richer and larger resources.

HPSG and LMF norm studies carried out so far suggest a method composed of two stages. The proposed method uses a projection rule system able to cover the different features that characterise lexical entries relative to the Arabic language.

The proposed method experimentation is intended to test the feasibility of the achieved system and to discern method limits. Evaluation of the prototype has been based on projection of Arabic HPSG lexicons that are constructed within the framework of several research works. These lexicons have varied structure and content that helped us to identify necessary conditions for the success of projection into LMF.

As for perspectives, we want to define the criteria allowing the formal verification of the projection success. Additionally, we want to try to supply applications conceived in unification grammars from lexical databases in conformity with LMF. This will hopefully encourage the reuse and the enrichment of existing lexicons. Finally, we can exploit projection of the LTAG grammars into HPSG while taking advantage of the phase that allows the conversion of canonical elementary trees into lexical entries that are specified in HPSG. This phase can be considered as an intermediate phase for the passage into LMF: from LTAG into HPSG and from HPSG into LMF.